\documentclass[review,3p,authoryear]{elsarticle}
\usepackage[multiple]{footmisc}
\usepackage{pdflscape}
\usepackage[dvipsnames]{xcolor}
\usepackage{amsmath}
\DeclareMathOperator*{\argmax}{arg\,max}

\usepackage{enumitem}
\usepackage{rotating} 
\usepackage{lineno}
\usepackage{amssymb}
\usepackage{sidecap}
\usepackage{graphicx}
\usepackage{caption}
\usepackage{subcaption}
\usepackage{multirow}
\usepackage{wrapfig}
\usepackage{xcolor}
\newcommand\crule[3][black]{\textcolor[HTML]{#1}{\rule{#2}{#3}}}
\usepackage{lipsum}
\usepackage{nicefrac}
\usepackage{ctable}
\usepackage{booktabs,tabularx}
\usepackage{adjustbox}
\usepackage{floatrow}
\usepackage{color}
\usepackage{xcolor}
\usepackage{colortbl}
\usepackage{float}
\restylefloat{table}
\newcolumntype{x}{>{\raggedright\arraybackslash}X}
\usepackage{bbding}
\usepackage{amsmath}
\usepackage{romannum}
\usepackage[colorlinks=true,
            linkcolor=magenta,
            urlcolor=gray,
            citecolor=cyan]{hyperref}
\usepackage{slashbox}
\usepackage{arydshln}
\newfloatcommand{capbtabbox}{table}[][\FBwidth]
\usepackage{blindtext}
\usepackage{enumitem}
\usepackage{algpseudocode}
\usepackage{xcolor}
\usepackage{adjustbox}
\usepackage{array}
\usepackage[linesnumbered,ruled,vlined]{algorithm2e}
\usepackage[final]{pdfpages}
\usepackage{amssymb}
\usepackage{pifont}
\newcommand{\cmark}{\ding{51}}%
\newcommand{\xmark}{\ding{55}}%
\usepackage{multicol}   
\usepackage{lipsum} 
\usepackage{xcolor}

\newcounter{includepdfpage}
\newcounter{currentpagecounter}

\newcolumntype{t}{>{\hsize=0.25\hsize}x}
\newcolumntype{s}{>{\hsize=0.75\hsize}x}
\newcolumntype{X}{>{\hsize=1.5\hsize}x}
\newcolumntype{R}[2]{%
    >{\adjustbox{angle=#1,lap=\width-(#2)}\bgroup}%
    l%
    <{\egroup}%
}

\newcommand{\tabincell}[2]{\begin{tabular}{@{}#1@{}}#2\end{tabular}}

\usepackage{xparse}
\NewDocumentCommand{\rot}{O{90} O{1em} m}{\makebox[#2][l]{\rotatebox{#1}{#3}}}%

\modulolinenumbers[5]
\newpageafter{author}
\begin{document}
\begin{frontmatter}


\title{The Liver Tumor Segmentation Benchmark (LiTS)}


\author[tum]{Patrick Bilic\textsuperscript{\textdagger$\ast$}}
\author[tum]{Patrick Christ\textsuperscript{\textdagger$\ast$}}
\author[tum,uzh]{Hongwei~Bran~Li\textsuperscript{$\ast$}}
\author[EPdeMotreal]{Eugene Vorontsov\textsuperscript{\textdagger$\ast$}}



\author[Tel-Aviv]{Avi Ben-Cohen\textsuperscript{\textdagger}}
\author[tumaim,tumdi,icluk]{Georgios~Kaissis\textsuperscript{\textdagger}}
\author[Jerusalem]{Adi Szeskin\textsuperscript{\textdagger}}

\author[Radboud]{Colin Jacobs\textsuperscript{\textdagger}}
\author[Radboud]{Gabriel Efrain~Humpire~Mamani\textsuperscript{\textdagger}}
\author[CRCHUM]{Gabriel~Chartrand\textsuperscript{\textdagger}}
% ...




\address[tum]{Department of Informatics, Technical University of Munich, Germany.}
\address[uzh]{Department of Quantitative Biomedicine, University of Zurich, Switzerland.}
\address[EPdeMotreal]{Ecole Polytechnique de Montr\'eal, Canada}
\address[Radboud]{Department of Medical Imaging, Radboud University Medical Center, Nijmegen, the Netherlands}
\address[Tel-Aviv]{Department of Biomedical Engineering, Tel-Aviv University, Israel}
% ...
\address[tumaim]{Institute for AI in Medicine, Technical University of Munich, Germany}
% ...
\address[tumdi]{Institute for diagnostic and interventional radiology, Klinikum rechts der Isar, Technical University of Munich, Germany}
% ...
\address[icluk]{Department of Computing, Imperial College London, London, United Kingdom}
% ...
\address[Jerusalem]{School of Computer Science and Engineering, the Hebrew University of Jerusalem, Israel}
% ...
\address[CRCHUM]{The University of Montr\'eal Hospital Research Centre (CRCHUM) Montr\'eal, Qu\'ebec, Canada}
\address[CRCHUM_an_tang]{Department of Radiology, Radiation Oncology and Nuclear Medicine, University of Montr\'eal, Canada}

% ...
\address[Radboud_dia]{Diagnostic Image Analysis Group, Radboud University Medical Center, Nijmegen, The Netherlands}
% ...


\let\thefootnote\relax\footnotetext{\textsuperscript{\textdagger} Organization team and data contributor.}

\let\thefootnote\relax\footnotetext{\textsuperscript{$\ast$} Patrick Bilic, Patrick Christ, Hongwei Bran Li, and Eugene Vorontsov made equal contributions to this work.}

\begin{abstract}
In this work, we report the set-up and results of the Liver Tumor Segmentation Benchmark (LiTS), which was organized in conjunction with the IEEE International Symposium on Biomedical Imaging (ISBI) 2017 and the International Conferences on Medical Image Computing and Computer-Assisted Intervention (MICCAI) 2017 and 2018. The image dataset is diverse and contains primary and secondary tumors with varied sizes and appearances with various lesion-to-background levels (hyper-/hypo-dense), created in collaboration with seven hospitals and research institutions. Seventy-five submitted liver and liver tumor segmentation algorithms were trained on a set of 131 computed tomography (CT) volumes and were tested on 70 unseen test images acquired from different patients. We found that not a single algorithm performed best for both liver and liver tumors in the three events. The best liver segmentation algorithm achieved a Dice score of 0.963, whereas, for tumor segmentation, the best algorithms achieved Dices scores of 0.674 (ISBI 2017), 0.702 (MICCAI 2017), and 0.739 (MICCAI 2018). Retrospectively, we performed additional analysis on liver tumor detection and revealed that not all top-performing segmentation algorithms worked well for tumor detection. The best liver tumor detection method achieved a lesion-wise recall of 0.458 (ISBI 2017), 0.515 (MICCAI 2017), and 0.554 (MICCAI 2018), indicating the need for further research.  
LiTS remains an active benchmark and resource for research, e.g., contributing the liver-related segmentation tasks in \url{http://medicaldecathlon.com/}. In addition, both data and online evaluation are accessible via \url{www.lits-challenge.com}.
\end{abstract}


\end{frontmatter}

\begin{center}
\noindent
\colorbox{YellowGreen}{\parbox{15cm}
{This is a pre-print of the journal article published in \emph{Medical Image Analysis}. If you wish to cite this work, please cite its journal version available here:  \url{https://doi.org/10.1016/j.media.2022.102680}. This work is available under CC-BY-NC-ND license.}} 
\end{center}

\section{Introduction}

\paragraph{Background} The liver is the largest solid organ in the human body and plays an essential role in metabolism and digestion. Worldwide, primary liver cancer is the second most common fatal cancer \citep{Stewart2014}.
Computed tomography (CT) is a widely used imaging tool to assess liver morphology, texture, and focal lesions \citep{hann2000}. 
Anomalies in the liver are essential biomarkers for initial disease diagnosis and assessment in both primary and secondary hepatic tumor disease \citep{Heimann}.
The liver is a site for primary tumors that start in the liver. In addition, cancer originating from other abdominal organs, such as the colon, rectum, and pancreas, and distant organs, such as the breast and lung, often metastasize to the liver during disease. Therefore, the liver and its lesions are routinely analyzed for comprehensive tumor staging. The standard Response Evaluation Criteria in Solid Tumor (RECIST) or modified RECIST protocols require measuring the diameter of the largest target lesion \citep{eisenhauer2009new}.
Hence, accurate and precise segmentation of focal lesions is required for cancer diagnosis, treatment planning, and monitoring of the treatment response.
Specifically, localizing the tumor lesions in a given image scan is a prerequisite for many treatment options such as thermal percutaneous ablation \citep{shiina2018percutaneous}, 
radiotherapy, surgical resection \citep{albain2009} and arterial embolization \citep{virdis2019clinical}.
Like many other medical imaging applications, manual delineation of the target lesion in 3D CT scans is time-consuming, poorly reproducible \citep{todorov2020} and segmentation shows operator-dependent results.

\begin{figure}[t]

\centering
         \includegraphics[width=\textwidth]{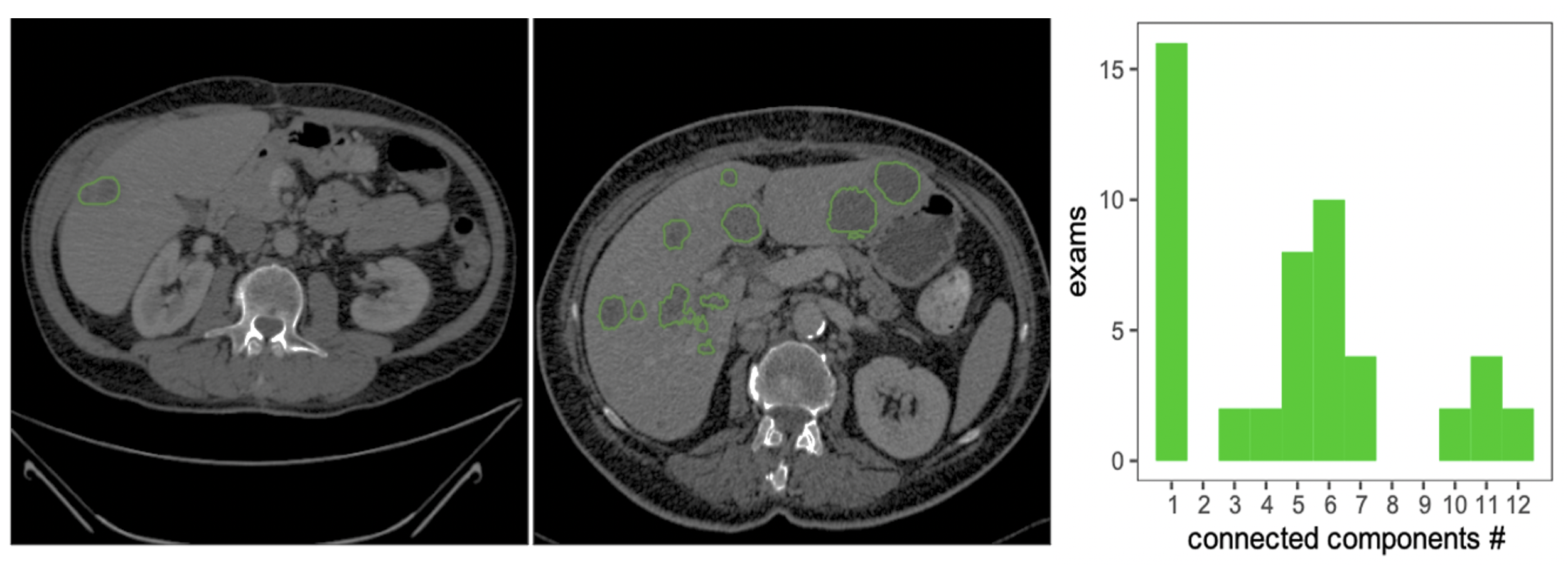}

\hfill
    \caption{Example from the LiTS dataset depicting a variety of shapes of on contrast-enhanced abdominal CT scans acquired. While most exams in the dataset contain only one lesion, a large group of patients with some (2-7) or many (10-12) lesions, as shown in the histogram calculated over the whole dataset.}
    \label{fig:motivation}
\end{figure}

\paragraph{Technical challenges} Fully automated segmentation of the liver and its lesions remain challenging in many aspects. First, the variations in the lesion-to-background contrast \citep{Moghbel2017} can be caused by: a) varied contrast agents, b) variations in contrast enhancement due to different injection timing, and c) different acquisition parameters (e.g., resolution, mAs and kVp exposure, reconstruction kernels). Second, the coexistence of different types of focal lesions (benign \emph{vs.} malignant and tumor sub-types) with varying image appearances presents an additional challenge for automated lesion segmentation. Third, the liver tissue background signal can vary substantially in the presence of chronic liver disease, which is a common precursor of liver cancer. It is observed that many algorithms struggle with disease-specific variability, including the differences in size, shape, and the number of lesions, as well as with modifications in shape and appearance to the liver organ itself induced by treatment \citep{Moghbel2017}. Examples of differences in liver and tumor appearance in two patients are depicted in Figure~\ref{fig:motivation}, demonstrating the challenges of generalizing to unseen test cases with varying lesions. 

\paragraph{Contributions} In order to evaluate the state-of-the-art methods for automated liver and liver tumor segmentation, we organized the Liver Tumor Segmentation Challenge (LiTS) in three events: 1) in conjunction with the IEEE International Symposium on Biomedical Imaging (ISBI) 2017, 2) with MICCAI 2017 and 3) as a dedicated challenge task on liver and liver tumor segmentation in the Medical Segmentation Decathlon 2018 in MICCAI 2018 \citep{antonelli2022medical}. 

In this paper, we describe the three key contributions to fully automated liver and liver tumor segmentation. First, we generate a new public multi-center dataset of 201 abdominal CT Volumes and the reference segmentations of liver and liver tumors. Second, we present the set-up and the summary of our LiTS benchmarks in three grand challenges. Third, we review, evaluate, rank, and analyze the resulting state-of-the-art algorithms and results. 
The paper is structured as follows: \hyperref[sec:priorwork]{Section II} reviews existing public datasets and state-of-the-art automated liver and liver tumors segmentation. Next, \hyperref[sec:method]{Section III} describes the LiTS challenge setup, the released multi-center datasets, and the evaluation process. \hyperref[sec:results]{Section IV} reports results, analyzes the liver tumor detection task, showcases critical cases in the LiTS Challenge results, and discusses the technical trends and challenges in liver tumor segmentation.  \hyperref[sec:discussion]{Section V} discusses the limitations, summarizes this work, and points to future work.

\begin{table}[h]
\scriptsize
\resizebox{\textwidth}{!}{
\begin{tabular}{llcccllllc}
\toprule
Dataset& Institution & Liver& Tumor  & Segmentation & \#Volumes & Modality \\ 
\midrule
TCGA-LIHC \citep{TCGALIHC0:online} &TCIA & \cmark & \cmark & \textcolor{orange}{\xmark} & 1688 & CT, MR, PT \\
MIDAS \citep{MIDASOri39:online} & IMAR & \cmark & \cmark & \textcolor{orange}{\xmark}  & 4& CT \\
3Dircadb-01 and 3Dircadb-02 \citep{soler20103d} & IRCAD & \cmark & \cmark & \cmark & 22 & CT \\
SLIVER’07 \citep{Heimann} & DKFZ & \cmark & \textcolor{orange}{\xmark} & \cmark & 30 & CT \\
LTSC’08 \citep{Heimann} & Siemens & \textcolor{orange}{\xmark} & \cmark & \cmark & 30 & CT \\
ImageCLEF’15 (\url{imageclef.org/2015}) & Bogazici Uni. & \cmark & \textcolor{orange}{\xmark} & \cmark & 30 & CT \\
VISCERAL’16 \citep{jimenez2016cloud} & Uni. of Geneva & \cmark & \textcolor{orange}{\xmark} & \cmark & 60/60 & CT/MRI \\
CHAOS’19 \citep{kavur2021chaos} & Dokuz Eylul Uni. & \cmark & \textcolor{orange}{\xmark} & \cmark & 40/120 & CT/MRI \\
\textbf{LiTS}       & \textbf{TUM}        & \cmark & \cmark & \cmark    & \textbf{201}                  & CT \\
\bottomrule
\end{tabular}}
\caption{Overview of publicly available medical datasets of liver and liver tumor images. The LiTS dataset offers a comparably large amount of 3D scans, including liver and liver tumor annotations.}
\label{tab:datasets}
\end{table}

\section{Prior Work: Datasets \& Approaches}
\label{sec:priorwork}
\color{black}

\subsection{Publicly available liver and liver tumor datasets.} 

Compared to other organs, available liver datasets offer either a relatively small number of images and reference segmentation or provide no reference segmentation (see Table~\ref{tab:datasets}). The first grand segmentation challenge - {SLIVER07} was held in MICCAI 2007 \citep{Heimann}, including the 30 CT liver images for automated segmentation.
In MICCAI 2008, the  {LTSC’08} segmentation challenge offered 30 CT volumes with a focus on tumor segmentation \citep{deng20083d}. 
The {ImageCLEF} 2015 liver CT reporting benchmark \footnote{\url{https://www.imageclef.org/2015/liver}} made 50 volumes available for computer-aided structured reporting instead of segmentation. 
The {VISCERAL} \citep{jimenez2016cloud} challenge provided 60 scans per two modalities (MRI and CT) for anatomical structure segmentation and landmark detection. The recent CHAOS challenge \citep{kavur2021chaos} provide 40 CT volumes and 120 MRI volumes for \emph{healthy} abdominal organ segmentation. However, none of these datasets represents well-defined cohorts of patients with lesions, and segmentation of the liver and its lesions are absent.

\subsection{Approaches for liver and liver tumour segmentation} 

Before 2016, most automated liver and tumor segmentation methods used traditional machine learning methods. However, since 2016 and the first related publications at MICCAI \citep{Christ2016}, deep learning methods have gradually become a methodology of choice. The following section provides an overview of published automated liver and liver tumor segmentation methods. 

\subsubsection{Liver segmentation}

Published work on liver segmentation methods can be grouped into three categories based on: (1) prior shape and geometric knowledge, (2) intensity distribution and spatial context, and (3) deep learning. 

\paragraph{Methods based on shape and geometric prior knowledge}
Over the last two decades, statistical shape models (SSMs) \citep{cootes1995} have been used for automated liver segmentation tasks. However, deformation limitations prevent SSMs from capturing the high variability of the liver shapes. To overcome this issue, SSM approaches often rely on additional steps to obtain a finer segmentation contour. Therefore SSMs followed by a deformable model performing free form deformation became a valuable method for liver segmentation \citep{heimann2006active,kainmuller2007shape,zhang2010automatic,tomoshige2014conditional,wang2015adaptivemesh}. Moreover, variations and further enhancement of SSMs such as 3D-SSM based on an intensity profiled appearance model \citep{lamecker2004segmentation}, incorporating non-rigid template matching \citep{saddi2007global}, initialization of SSMs utilizing an evolutionary algorithm \citep{heimann2007shape}, hierarchical SSMs \citep{ling2008hierarchical}, and deformable SSMs \citep{zhang2010automatic} had been proposed to solve liver segmentation tasks automatically. SSMs-based methods showed the best results in {SLIVER07}, the first grand challenge held in MICCAI 2007 \citep{Heimann,dawant2007semi}.

\paragraph{Methods based on intensity distribution and spatial context} A probabilistic atlas (PA) is an anatomical atlas with parameters that are learned from a training dataset. Park \emph{et al.} proposed the first PA utilizing 32 abdominal CT series for registration based on mutual information and thin-plate splines as warping transformations \citep{park2003PA} and a Markov random field (MRF) \citep{park2003PA} for segmentation. Further proposed atlas-based methods differ in their computation of the PA and how the PA is incorporated into the segmentation task. Furthermore, PA can incorporate relations between adjacent abdominal structures to define an anatomical structure surrounding the liver \citep{zhou2006PA}. Multi-atlas methods improved liver segmentation results by using non-rigid registration with a B-spline transformation model \citep{slagmolen2007atlas}, dynamic atlas selection and label fusion \citep{xu2015efficient}, or liver and non-liver voxel classification based on k-Nearest Neighbors \citep{van2007automatic}.

Graph cut methods offer an efficient way to binary segmentation problems, initialized by adaptive thresholding \citep{massoptier2007fully} and supervoxel \citep{wu2016automatic}. 

\paragraph{Methods based on deep learning}
In contrast to the methods above, deep learning, especially convolutional neural networks (CNN), is a data-driven method that can be optimized end-to-end without hand-craft feature engineering \citep{litjens2017survey}. 
The U-shape CNN architecture \citep{Unet} and its variants \citep{milletari2016v,isensee2020nnu,li2018fully} are widely used for biomedical image segmentation and have already proven their efficiency and robustness in a wide range of segmentation tasks. 
Top-performing methods share the commonality of multi-stage processes, beginning with a 3D CNN for segmentation, and post-process the resulting probability maps with Markov random field \citep{dou20163d}. 
Many early deep learning algorithms for liver segmentation combine neural networks with dedicated post-processing routines: \cite{Christ2016} uses 3D fully connected neural networks combined with conditional random fields, \cite{hu2016automatic} rely on a 3D CNN followed by a surface model. In contrast, \cite{lu2017automatic} use a CNN regularized by a subsequent graph-cut segmentation. 

\subsubsection{Liver tumor segmentation}
Compared to the liver, its lesions feature a more comprehensive range of shape, size, and contrast. Liver tumors can be found in almost any location, often with ambiguous boundaries. Differences in the uptake of contrast agents may introduce additional variability. Therefore liver tumor segmentation is considered to be the more challenging task. Published methods of liver tumor segmentation can be categorized into 1) thresholding and spatial regularization, 2) local features and learning algorithms, and 3) deep learning. 

\paragraph{Methods with thresholding and spatial regularization}
Based on the assumption that gray level values of tumor areas differ from pixels/voxels belonging to regions outside the tumor, thresholding is a simple yet effective tool to automatically separate tumor from liver and background, first shown by Soler \emph{et al.} \citep{soler2001fully}. Since then the threshold have set by histogram analysis \citep{ciecholewski2007automatic}, maximum variance between classes \citep{nugroho2008contrast} and iterative algorithm \citep{abdel2010fully} to improve tumor segmentation results. Spatial regulation techniques rely on (prior) information about the image or morphologies, e.g., tumor size, shape, surface, or spatial information. This knowledge is used to introduce constraints in the form of regularization or penalization. Adaptive thresholding methods can be combined with model-based morphological processing for heterogeneous lesion segmentation \citep{moltz2008segmentation,moltz2009advanced}. Active contour \citep{kass1988snakes} based tumor segmentation relies on shape and surface information and utilize probabilistic models \citep{ben2008liver} or histogram analysis \citep{linguraru2012tumor} to create segmentation maps automatically. Level set \citep{osher1988fronts} methods allow numerical computations of tumor shapes without parametrization. Level set approaches for liver tumor segmentation are combined with supervised pixel/voxel classification in 2D \citep{smeets2008segmentation} and 3D \citep{jimenez2011optimal}.

\label{ss:tumorsegclassification}
\paragraph{Methods using local features and learning algorithms}
Clustering methods include k-means \citep{massoptier2008new} and fuzzy c-means clustering with segmentation refinement using deformable models \citep{hame2008liver}. Among supervised classification methods are a fuzzy classification based level set approach \citep{smeets2008segmentation}, support vector machines in combination with a texture based deformable surface model for segmentation refinement \citep{vorontsov2014metastatic}, AdaBoost trained on texture features \citep{shimizu2008ensemble} and image intensity profiles \citep{li2006machine}, logistic regression \citep{wen2009comparison}, and random forests recursively classifying and decomposing supervoxels \citep{conze2017scale}. 

\paragraph{Methods based on deep learning}
Before LiTS, deep learning methods have been rarely used for liver tumor segmentation tasks. \cite{Christ2016} was the first to use 3D U-Net for liver and liver tumor segmentation, proposing a cascaded segmentation strategy, together with a 3D conditional random field refinement. Many of the subsequent deep learning approaches were developed and tested in conjunction with the LiTS dataset.

Benefiting from the availability of the LiTS public dataset, many new deep learning solutions on liver and liver segmentation were proposed. U-Net-based architectures are extensively used and modified to improve segmentation performance. For example, the nn-UNet \citep{isensee2020nnu} first presented in LiTS at MICCAI 2018, was shown to be one of the most top-performing methods in 3D image segmentation tasks. 
The related works will be discussed in the results section.

\section{Methods} \label{sec:method}
\subsection{Challenge setup} \label{sec:setup}
The first LiTS benchmark was organized in Melbourne, Australia, on April 18, 2017, in a workshop held at the IEEE ISBI 2017 conference. During Winter 2016/2017, participants were solicited through private emails, public email lists, social media, and the IEEE ISBI workshop announcements. 
Participants were requested to register at our online benchmarking system hosted on CodaLab and could download annotated training data and unannotated test data. The online benchmarking platform automatically computed performance scores. They were asked to submit a four-page summary of their algorithm after successful submissions to the CodaLab platform. Following the successful submission process at ISBI 2017, the second LiTS benchmark was held on September 14, 2017, in Quebec City, Canada, as a MICCAI workshop. The third edition of LiTS was a part of the Medical Segmentation Decathlon at MICCAI 2018 (available at \url{http://medicaldecathlon.com/}). 

At ISBI 2017, five out of seventeen participating teams presented their methods at the workshop. 
At MICCAI 2017, the LiTS challenge introduced a new benchmark task - liver segmentation. Participants registered a new CodaLab benchmark and were asked to describe their algorithm after the submission deadline, resulting in 26 teams. The training and test data for the benchmark were identical to the ISBI benchmark. The workshop at the MICCAI 2017 was organized similarly to the ISBI edition.
At MICCAI 2018, LiTS was a part of a medical image segmentation decathlon organized by King's College London in conjunction with eleven partnerships for data donation, challenge design, and administration. The LiTS benchmark dataset described in this paper constitutes the decathlon's liver and liver lesion segmentation tasks. However, the overall challenge also required the participants to address nine other tasks, including brain tumor, heart, hippocampus, lung, pancreas, prostate, hepatic vessel, spleen, and colon segmentation. To this end, algorithms were not necessarily optimized only for liver CT segmentation. 

\subsection{Dataset}
Training and test cases both represented abdomen CT images. The data is licensed as CC BY-NC-SA. Only the organizers from TUM have access to the labels of test images. The participants could download annotated training data from the LiTS Challenge website \footnote{\url{www.lits-challenge.com}}.

\paragraph{Contributors} The image data for the LiTS challenge are collected from seven clinical sites all over the world, including a) Rechts der Isar Hospital, the Technical University of Munich in Germany, b) Radboud University Medical Center, the Netherlands, c) Polytechnique Montréal and CHUM Research Center in Canada, d) Sheba Medical Center in Israel, e) the Hebrew University of Jerusalem in Israel, f) Hadassah University Medical Center in Israel, and g) IRCAD in France. The distribution of the number of scans per institution is described in Table \ref{tab:datasetnumbers}. The LiTS benchmark dataset contains 201 computed tomography images of the abdomen, of which 194 CT scans contain lesions. All data are anonymized, and the images have been reviewed visually to preclude the presence of personal identifiers. The only processing applied to the images is a transformation into a unified NIFTY format using NiBabel in Python \footnote{\url{https://nipy.org/nibabel/}}. All parties agreed to make the data publicly available; ethics approval was not required.

\begin{table}[t]
\scriptsize
\begin{tabular}{l|cc}
\toprule
\multicolumn{1}{c}{Institutions} & Train       & Test        \\ \hline
Rechts der Isar Hospital, TUM, Germany                                                                 & 28             & 28         \\ \hline
Radboud University Medical Center, the Netherlands                                                    & 48             & 12         \\ \hline
Polytechnique Montreal and CHUM Research Center in Canada                     & 30             & 25         \\ \hline 
IRCAD, France    & 20             & 0          \\ \hline
Sheba Medical Center &      &  \\ \cline{1-1}
Hebrew University of Jerusalem  &   5   &  5  \\ \cline{1-1} 
Hadassah University &              &           \\ \hline 

Total                                                                        & 131            & 70         \\ \bottomrule
\end{tabular}
\caption{Distribution of the number of scans per institution in the train and test in the LiTS dataset.}
\label{tab:datasetnumbers}
\end{table}


\begin{table}[]
\scriptsize
\begin{tabular}{lccl}
\\ \toprule
\multicolumn{1}{c}{} & Train & Test & p-value \\ \midrule
In-plane resolution (mm) & 0.76 (0.7, 0.85)  & 0.74 (0.69, 0.8)   & 0.042   \\ \hline
Slice thickness (mm)     & 1.0 (0.8, 1.5)                                     & 1.5 (0.8, 4.0)                                   & 0.004   \\ \hline
Volume size                     & \begin{tabular}[c]{@{}c@{}}512$\times$512$\times$432\\  (512$\times$512$\times$190 , 512$\times$512$\times$684)\end{tabular} & \begin{tabular}[c]{@{}c@{}}512$\times$512$\times$270\\  (512$\times$512$\times$125 , 512$\times$512$\times$622)\end{tabular} & 0.058   \\ \hline
Number of tumors     & 3~(1, 9)                                       & 5~(2, 12)         & 0.016   \\ \hline
Tumor volume (mm\textsuperscript{3}) & \begin{tabular}[c]{@{}c@{}}$16.11\times10^{3}$ 
\\ $(3.40\times10^{3}$, $107.77\times10^{3}$) \end{tabular}                     & \begin{tabular}[c]{@{}c@{}}$34.78\times10^{3}$\\ ($10.41\times 10^{3},98.90\times10^{3}$)\end{tabular}                        & 0.039   \\ \hline
Liver volume (mm\textsuperscript{3})             & \begin{tabular}[c]{@{}c@{}}$1586.48\times10^{3}\pm447.14\times10^{3}$\\ ($1337.75\times10^{3},1832.46\times10^{3}$)\end{tabular}                  & \begin{tabular}[c]{@{}c@{}}$1622.64\times10^{3}\pm546.48\times10^{3}$\\ ($1315.94\times10^{3},1977.27\times10^{3}$)\end{tabular}                  & 0.60    \\ \hline
\end{tabular}
\caption{The characteristics of the LiTS training and test sets. The median values and the interquartile range (IQR) are shown for each parameter. In addition, P-values were obtained by Mann-whitney u test, describing the significance between the training set and test set shown in the last column.
An alpha level of 0.05 was chosen to determine significance.}
\label{tab:datasetstats}
\end{table}

\paragraph{Data diversity}
The studied cohort covers diverse types of liver tumor diseases, including primary tumor disease (such as hepatocellular carcinoma and cholangiocarcinoma) and secondary liver tumors (such as metastases from colorectal, breast and lung primary cancers). The tumors had varying lesion-to-background ratios (hyper- or hypo-dense). The images represented a mixture of pre- and post-therapy abdominal CT scans and were acquired with different CT scanners and acquisition protocols, including imaging artifacts (e.g., metal artifacts) commonly found in real-world clinical data. Therefore, it was considered to be very diverse concerning resolution and image quality. The in-plane image resolution ranges from 0.56 mm to 1.0 mm, and 0.45 mm to 6.0 mm in slice thickness. Also, the number of axial slices ranges from 42 to 1026. The number of tumors varies between 0 and 12. The size of the tumors varies between 38 mm\textsuperscript{3} and 1231 mm\textsuperscript{3}. The test set shows a higher number of tumor occurrences compared to the training set. The statistical test ({p}-value=0.6) shows that the liver volumes in the training and test sets do not differ significantly. The average tumor HU value is 65 and 59 in the train and test sets, respectively. 
The LiTS data statistics are summarized in Table~\ref{tab:datasetstats}. 
The training and test split is with a ratio of 2:1 and the training and test sets were similar in center distribution. Generalizability to unseen centers has, hence, not been tested in LiTS.
\paragraph{Annotation protocol} The image datasets were annotated manually using the following strategy: A radiologist with $>$3 years of experience in oncologic imaging manually labelled the datasets slice-wise using the ITK-SNAP \citep{yushkevich2006user} software and assigning one of the labels 'Tumor' or 'Healthy Liver'. Here, the “Tumor” label included any neoplastic lesion irrespective of origin (i.e. both primary liver tumors and metastatic lesions). Any part of the image not assigned one of the aforementioned labels was considered 'Background'. The segmentations were verified by three further readers blinded to the initial segmentation, with the most senior reader serving as tie-breaker in cases of labelling conflicts. Those scans with very small and uncertain lesion-like structures were omitted in the annotation.

\subsection{Evaluation}
\subsubsection{Ranking strategy} The main objective of \emph{LiTS} was to benchmark segmentation algorithms. We assessed the segmentation performance of the LiTS submissions considering three aspects: a) volumetric overlap, b) surface distance, and c) volume similarity. All the values of these metrics are released to the participating teams. Considering that the volumetric overlap is our primary interest in liver and liver tumor segmentation, for simplicity, we only use the Dice score to rank the submissions at ISBI-2017 and MICCAI-2017. However, the exact choice of evaluation metric does sometimes affect the ranking results, as different metrics are sensitive to different types of segmentation errors. Hence, we provide a post-challenge ranking, considering three properties by summing up three ranking scores and re-ranking by the final scores. The evaluation codes for all metrics can be accessed in Github \footnote{\url{https://github.com/PatrickChrist/lits-challenge-scoring}}.

\subsubsection{Statistical tests} To compare the submissions from two teams in a \emph{per case} manner, we used Wilcoxon signed-rank test \citep{rey2011wilcoxon}. To compare the distributions of submissions from two years, we used the Mann-Whitney U Test \citep{mcknight2010mann} (unpaired) for the two groups.

\subsubsection{Segmentation metrics}

\paragraph{Dice score}
\label{sec:Dice_score}
The Dice score evaluates the degree of overlap between the predicted and reference segmentation masks. For example, given two binary masks $A$ and $B$, it is formulated as:

\begin{equation}\label{eq:dice}
Dice(A,B) = \frac{2| A \cap B |}{|A| + |B|}
\end{equation}

The Dice score is applied \emph{per case} and then averaged over all cases consistently for three benchmarks. This way, the Dice score applies a higher penalty to prediction errors in cases with fewer actual lesions.



\paragraph{Average symmetric surface distance}
\label{sec:ASD}



Surface distance metrics are correlated measures of the distance between the surfaces of a reference and the predicted region. Let $S(A)$ denote the set of surface voxels of $A$. Then, the shortest distance of an arbitrary voxel $v$ to $S(A)$ is defined as:
\begin{equation}\label{eq:dist}
            d(v,S(A)) = \min_{s_A\in S(A)} ||v-s_A||,
\end{equation}

where $||.||$ denotes the Euclidean distance. The average symmetric surface distance ({ASD}) is then given by:
\begin{equation}\label{eq:asd}
            ASD(A,B) =
                \frac{1}{|S(A)|+|S(B)|}
                \left(
                    \sum_{s_A\in S(A)} d(s_A,S(B))
                    +
                    \sum_{s_B\in S(B)} d(s_B,S(A))
                \right).
\end{equation}

\paragraph{Maximum symmetric surface distance}
The maximum symmetric surface distance ({MSSD}), also known as the Symmetric Hausdorff Distance, is similar to ASD except that the maximum distance is taken instead of the average:

\begin{equation}\label{eq:msd}
MSSD(A,B) = \max
                \left\{
                    \max_{s_A\in S(A)} d(s_A,S(B)),
                    \max_{s_B\in S(B)} d(s_B,S(A))
                \right\}.
\end{equation}

\paragraph{Relative volume difference}
The relative volume difference (RVD) directly measures the volume difference without considering the overlap between reference $A$ and the prediction $B$. 
\begin{equation}\label{eq:rvd}
RVD(A,B) = \frac{|B| - |A|}{|A|}.
\end{equation}

For the other evaluation metrics, such as tumor burden estimation and the corresponding rankings, please check the Appendix. 

\subsubsection{Detection metrics}
Considering the clinical relevance of lesion detection, we introduce three detection metrics in the additional analysis. The metrics are calculated globally to avoid potential issues when the patient has no tumors. There must be a known correspondence between predicted and reference lesions to evaluate the lesion-wise metrics. Since \emph{lesions} are all defined as a single binary map, this correspondence must be determined between the prediction and reference masks' connected components. Components may not necessarily have a one-to-one correspondence between the two masks. The details of the correspondence algorithm are presented in \ref{correspondence_alg}. 

\emph{Individual lesions} are defined as 3D connected components within an image. A lesion is considered detected if the predicted lesion has sufficient overlap with its corresponding reference lesion, measured as the intersection over the union of their respective segmentation masks. It allows for a count of true positive, false positive, and false-negative detection, from which we compute the \emph{precision} and \emph{recall} of lesion detection. 
The metrics are defined as follows: 

\emph{Individual lesions} are defined as 3D connected components within an image. A lesion is considered detected if the predicted lesion has sufficient overlap with its corresponding reference lesion, measured as the intersection over the union of their respective segmentation masks. It allows for a count of true positive, false positive, and false-negative detection, from which we compute the \emph{precision} and \emph{recall} of lesion detection. 
The metrics are defined as follows: 

\begin{equation}\label{eq:IoU}
    IoU = \frac{| A \cap B |}{|A \cup B|}.
\end{equation}

\paragraph{Precision} It relates the number of true positives (TP) to false positives (FP), also known as positive predictive value:
\begin{equation}\label{eq:precision}
precision = {{TP} \over {TP+FP}}.
\end{equation}

\paragraph{Recall} It relates the number of true positives (TP) to false negatives (FN), also known as sensitivity or true positive rate:
\begin{equation}\label{eq:recall}
recall = {{TP} \over {TP+FN}}.
\end{equation}

\paragraph{F1 score} It measures the harmonic mean of precision and recall:
\begin{equation}\label{eq:f1}
    F1 = {{2} \over {precision^{-1}+ recall^{-1}}}.
\end{equation}





\subsubsection{Participating policy and online evaluation platform}
The participants were allowed to submit three times per day in the test stage during the challenge week. Members of the organizers' groups could participate but were not eligible for awards. The awards were given to the top three teams for each task. The top three performing methods gave 10-min presentations and were announced publicly. For a fair comparison, the participating teams were only allowed to use the released training data to optimize their methods. All participants were invited to be co-authors of the manuscript summarizing the challenge.

A central element of LiTS was - and remains to be - its online evaluation tool hosted by CodaLab. 
On Codalab, participants could download annotated training and "blinded" test data and upload their segmentation for the test cases. The system automatically evaluated the uploaded segmentation maps' performance and made the overall performance available to the participants. Average scores for the different liver and lesion segmentation tasks and tumor burden estimation were also reported online on a leaderboard.
Reference segmentation files for the LiTS test data were hosted on the Codalab but not accessible to participants. Therefore, the users uploaded their segmentation results through a web interface, reviewed the uploaded segmentation, and then started an automated evaluation process. 
The evaluation to assess the segmentation quality took approximately two minutes per volume. In addition, the overall segmentation results of the evaluation were automatically published on the Codalab leaderboard web page and could be downloaded as a csv file for further statistical analysis. 

The Codalab platform remained open for further use after the three challenges and will remain so in the future. As of April 2022, it has evaluated more than 3,414 valid submissions (238,980 volumetric segmentation) and recorded over 900 registered LiTS users. 
 The up-to-date ranking is available at Codalab for researchers to continuously monitor new developments and streamline improvements. In addition, the code to generate the evaluation metrics between reference and predictions is available as open-source at GitHub \footnote{\url{https://github.com/PatrickChrist/LiTS-CHALLENGE}}.


\section{Results}
\label{sec:results}
\subsection{Submitted algorithms and method description}
The submitted methods in ISBI-2017, MICCAI-2017 and MICCAI-2018 are summarized in Table \ref{tab:method_1} and Table \ref{tab:method_2}, and the reference paper of Medical Segmentation Decathlon \citep{antonelli2022medical}. In the following, we grouped the algorithms with several properties. 
\paragraph{Algorithms and architectures} Seventy-three submissions were fully automated approaches, while only one was semi-supervised (\crule[b000b0]{0.20cm}{0.20cm} J. Ma \emph{et al.}). 
U-Net derived architectures were overwhelmingly used in the challenge with only two automated methods using a modified VGG-net (\crule[00f0ff]{0.20cm}{0.20cm} J. Qi \emph{et al.}) and a k-CNN (\crule[ff00ff]{0.20cm}{0.20cm} J. Lipkova \emph{et al.}) respectively. 
Most submissions adopted the coarse-to-fine approach in which multiple U-nets were cascaded to perform liver and liver segmentation at different stages. Additional residual connections and adjusted input resolution were the most common changes to the basic U-Net architecture. Three submissions combined individual models as an ensemble technique. In 2017, 3D methods were not directly employed on the original image resolution by any of the submitted methods due to high computational complexity. However, some submissions used 3D convolutional neural networks solely for tumor segmentation tasks with small input patches. Instead of full 3D, other methods tried to capture the advantages of three-dimensionality by using a 2.5 D model architecture, i.e., providing a stack of images as a multi-channel input to the network and receiving the segmentation mask of the center slice of this stack as a network output.

\paragraph{Critical components of the segmentation methods} Data pre-processing with HU-value clipping, normalization, and standardization were the most frequent techniques in most of the methods. Data augmentation was also widely used and mainly focused on standard geometric transformations such as flipping, shifting, scaling, or rotation. Individual submissions implemented more advanced techniques such as histogram equalization and random contrast normalization. The most common optimizer varied between ADAM and Stochastic gradient descent with momentum, with one approach relying on RMSProp. Multiple loss functions were used for training, including standard and class-weighted cross-entropy, Dice loss, Jaccard loss, Tversky loss, L2 loss, and ensemble loss techniques combining multiple individual loss functions into one. 

\paragraph{Post-processing} Some types of post-processing methods were also used by the vast majority of the algorithm. The common post-processing steps were to form connected tumor components and overlay the liver mask on the tumor segmentation to discard tumors outside the liver region. More advanced methods included a random forest classifier, morphological filtering, a particular shallow neural network to eliminate false positives or custom algorithms for tumor hole filling. 

\paragraph{Features of top-performing methods} The best-performing methods at ISBI 2017 used cascaded U-Net approaches with short and long skip connections and 2.5D input images (\crule[ff0000]{0.20cm}{0.20cm} X. Han \emph{et al.}). In addition, weighted cross-entropy loss functions and a few ensemble learning techniques were employed by most of the top-performing methods, together with some common pre- and post-processing steps such as HU-value clipping and connected component labeling, respectively. Some top-performing submissions at MICCAI 2017 (e.g., \crule[baba00]{0.20cm}{0.20cm} J. Zou) integrated the insights from the ISBI 2017, including the idea of the ensemble, adding residual connections, and featuring more sophisticated rule-based post-processing or classical machine learning algorithms. Therefore, the main architectural differences compared to the ISBI submissions were the higher usage of ensemble learning methods, a higher incidence of residual connections, and an increased number of more sophisticated post-processing steps. 
Another top-performing method by~\crule[008900]{0.20cm}{0.20cm}~X. Li \emph{et al.} proposed a hybrid insight by integrating the advantages of the 2D and 3D networks in the 3D liver tumor segmentation task \citep{li2018h}. Therefore, compared to the methods in ISBI submissions that solely rely on 2D or 3D convolutions, the main architecture difference was the hybrid usage of 2D and 3D networks. 
In MICCAI-LiTS 2018, 3D deep learning models became popular and generally outperformed 2.5D or 2D without more sophisticated pre-processing steps. 




\begin{center}
\begin{landscape}
\begin{table}
\tiny
\caption{Details of the participating teams’ methods in LiTS-ISBI-2017.} 
\setlength\tabcolsep{4pt} 
\begin{tabular}{l l l c c c c c c c c c }
\toprule
\tabincell{l}{\textbf{Lead Author} \& \\ \textbf{Team Members}} & \tabincell{l}{\textbf{Method, Architecture} \& \\ \textbf{Modifications}} & \tabincell{l}{\textbf{Data Augmentation}} & \textbf{Loss Function}  &\textbf{Optimizer Training} & \textbf{Pre-processing}  &\textbf{Post-processing} & \textbf{Ensemble strategy}\\
\midrule
\crule[ff0000]{0.15cm}{0.15cm} \textbf{X. Han};  & \tabincell{l}{Residual U-Net with 2.5D \\ input (a stack of 5 slices)} & cropping, flipping &  weighted cross-entropy & SGD with momentum & \tabincell{l}{Value-clipping to  \\ $[$-200, 200$]$}& \tabincell{l}{Connected components with \\maximum probability below \\0.8 were removed} & None\\
\midrule
\tabincell{l}{\crule[00b0b0]{0.15cm}{0.15cm} \textbf{E. Vorontsov}; \\ A. Tang, C. Pal, \\ S. Kadoury}  & \tabincell{l}{Liver FCN provides pre-trained weights \\ for tumour FCN. Trained on 256$\times$256, \\ finetuned on 512$\times$512.}  & \tabincell{l}{Random flips, rotations, \\ zooming, elastic \\deformations.} & Dice loss & RMSprop & None & \tabincell{l}{largest connected \\component for liver}& Ensemble of three models\\

\midrule
\tabincell{l}{\crule[b0b0ff]{0.15cm}{0.15cm} \textbf{G. Chlebus}; \\ H. Meine, \\J. H. Moltz, \\ A. Schenk}  & \tabincell{l}{Liver: 3 orthogonal 2D U-nets working \\ on four resolution levels. \\ Tumor: 2D U-net working on four \\resolution levels}  &  \tabincell{l}{Mannual removal of cases \\with flawed reference \\segmentation}  & soft Dice & Adam & \tabincell{l}{Liver: resampling to \\ isotropic 2 mm voxels.} & \tabincell{l}{Tumor candidate filtering \\based on RF-classifier to \\ remove false positives} & \tabincell{l}{For liver: majority vote}\\
\midrule
\tabincell{l}{\crule[f3f00f]{0.15cm}{0.15cm} \textbf{L. Bi}; \\ Jinman Kim } & \tabincell{l}{Cascaded ResNet based on a \\ pre-trained FCN on 2D axial slices.} & \tabincell{l}{Random scaling, \\crops and flips} & cross-entropy & SGD & \tabincell{l}{Value-clipping to \\ $[$-160, 240$]$} &  \tabincell{l}{Morphological filter \\ to fill the holes}& \tabincell{l}{Multi-scale ensembling by \\averaging the outputs from. \\ different inputs sizes.}\\
\midrule
\crule[4949ff]{0.15cm}{0.15cm} \textbf{C. Wang} &  \tabincell{l}{Cascaded 2D U-Net in \\three orthogonal views} & \tabincell{l}{Random rotation, \\random translation, \\and scaling }& Soft Dice  & SGD  & None &  None & None\\
\midrule
 \crule[870000]{0.15cm}{0.15cm} \textbf{P. Christ} & \tabincell{l}{Cascaded U-Net} & \tabincell{l}{mirror, cropping,\\ additional noise} & weighted cross-entropy & SGD with momentum &   &  3D Conditional Random Field\\

\midrule
\tabincell{l}{\crule[ff00ff]{0.15cm}{0.15cm} \textbf{J. Lipkova}; \\ M. Rempfler, \\J. Lowengrub, \\ B. Menze }  & \tabincell{l}{ U-Net for liver segmentation\\ and Cahn-Hilliard Phase field \\ separation for lesions} & None & \tabincell{l}{Liver:cross-entropy; \\ Tumor:
Energy function} &  SGD & None & None & None \\
\midrule
\tabincell{l}{\crule[b000b0]{0.15cm}{0.15cm} \textbf{J. Ma}; \\
Y. Li, Y. Wu, \\
M. Zhang, X. Yang} & \tabincell{l}{Random Forest and \\Fuzzy Clustering} & None & None & None  &  \tabincell{l}{Value-clipping to  \\ $[$-160, 240$]$ and intensity \\ normalization to $[0,255$]}\\
\midrule
\tabincell{l}{\crule[870087]{0.15cm}{0.15cm} \textbf{T. Konopczynski}; \\ K. Roth, J. Hesser} &  \tabincell{l}{Dense 2D U-Nets (Tiramisu)}  & None &  \tabincell{l}{Soft Tversky-Coefficient \\ based Loss function}  &  Adam & \tabincell{l}{Value-clipping to \\ $[$-100, 400$]$} & None & None \\
\midrule
\tabincell{l}{\crule[b0b0ff]{0.15cm}{0.15cm} \textbf{M. Bellver}; \\ K. Maninis, J. Tuset, \\X. Giro-i-Nieto, \\ J. Torres}& \tabincell{l}{Cascaded FCN with side outputs at\\ different resolutions. \\Three-channel 2D input.} & None & \tabincell{l}{Weighted binary \\ cross entropy} & \tabincell{l}{SGD with\\ Momentum}  & \tabincell{l}{Value-clipping to  \\ $[$-150, 250$]$ and intensity \\ normalization to $[0,255$]}. & \tabincell{l}{Component analysis to \\ remove false positives. \\ 3D CRF.} & None\\
\midrule
\tabincell{l}{ \crule[00f0ff]{0.15cm}{0.15cm} \textbf{J. Qi}; \\M. Yue} & \tabincell{l}{A pretrained VGG with \\ concatenated multi-scale \\feature maps} & None  & binary cross entropy & SGD  & None & None & None \\

\bottomrule
\end{tabular}
\label{tab:method_1}
\end{table}

\end{landscape}
\end{center}


\begin{landscape}
\begin{table}
\tiny
\caption{Details of the participating teams’ methods in LiTS-MICCAI-2017.}  
\setlength\tabcolsep{4pt} 

\begin{tabular}{l l c c c c c c c c c c }
\toprule
\tabincell{l}{\textbf{Lead Author} \& \\ \textbf{Team Members}} & \tabincell{l}{\textbf{Method, Architecture} \& \\ \textbf{Modifications}} & \tabincell{l}{\textbf{Data Augmentation}} & \textbf{Loss Function}  &\textbf{Optimizer Training} & \textbf{Pre-processing}  &\textbf{Post-processing} & \textbf{Ensemble strategy}\\
\midrule
\crule[00b000]{0.15cm}{0.15cm} \textbf{Y. Yuan};  & \tabincell{l}{Hierarchical 2.5D \\ FCN network}  & \tabincell{l}{flipping, shifting, rotating, \\scaling and random contrast\\ normalization} & Jaccard distance  & Adam & \tabincell{l}{Clipping HU values \\ to $[$-100, 400$]$}& None & \tabincell{l}{ensembling 5 models \\from 5-fold cross validation} \\ [-2pt]
\midrule
\tabincell{l}{\crule[e4e400]{0.15cm}{0.15cm} \textbf{A. Ben-Cohen};}  & \tabincell{l}{VGG-16 as a backbone \\and 3-channel input} & scaling & softmax log loss &  SGD& \tabincell{l}{Clipping HU values \\ to $[$-160, 240$]$} &None & None \\ [-2pt]

\midrule
\tabincell{l}{\crule[baba00]{0.15cm}{0.15cm} \textbf{J. Zou}; } & Cascaded U-Nets   & weighted cross-entropy & Adam &  & \tabincell{l}{Clipping HU values \\ to $[$-75, 175$]$} & \tabincell{l}{hole filling and \\ noise removal} & \tabincell{l}{ensemble of two model \\ with different inputs} \\ [-2pt]
\midrule
\tabincell{l}{\crule[008900]{0.15cm}{0.15cm}  \textbf{X. Li}, H. Chen, \\ X. Qi, Q. Dou, \\ C. Fu, P. Heng  } & \tabincell{l}{H-DenseUNet \\ \citep{li2018h}} & \tabincell{l}{  rotation, flipping, scaling} & Cross-entropy & SGD & \tabincell{l}{Clipping HU values \\ to $[$-200, 250$]$} &\tabincell{l}{Largest connected \\ component; hole filling} & None \\ [-2pt]
\midrule
\tabincell{l}{\crule[545400]{0.15cm}{0.15cm} \textbf{G. Chlebus}; \\ H. Meine, \\J. H. Moltz, \\ A. Schenk}  & \tabincell{l}{Liver: 3 orthogonal 2D U-nets working \\ on four resolution levels. \\ Tumor: 2D U-net working on four \\resolution levels \citep{chlebus2018automatic}}  & \tabincell{l}{Mannual removal of cases \\with flawed reference \\segmentation}  & soft Dice & Adam & \tabincell{l}{Liver: resampling to \\ isotropic 2 mm voxels.} & \tabincell{l}{Tumor candidate filtering \\based on RF-classifier to \\ remove false positives} & \tabincell{l}{For liver: majority vote}\\  [-2pt]
\midrule
\crule[00ff00]{0.15cm}{0.15cm} \textbf{J. Wu} & \tabincell{l}{Cascade 2D FCN} & scaling &Dice Loss & Adam & None & None & None\\ 

\midrule
\crule[550500]{0.15cm}{0.15cm} \textbf{C. Wang} &  \tabincell{l}{Cascade 2D U-Net in \\three orthogonal views} & \tabincell{l}{Random rotation, \\random translation, \\and scaling }& Soft Dice  & SGD  & None &  None & None\\ [-2pt]

\midrule
\tabincell{l}{\crule[00b0b0]{0.15cm}{0.15cm} \textbf{E. Vorontsov}; \\ A. Tang, C. Pal, \\ S. Kadoury}  & \tabincell{l}{Liver FCN provides pre-trained weights \\ for tumour FCN. Trained on 256$\times$256, \\ finetuned on 512$\times$512.}  & \tabincell{l}{Random flips, rotations, \\ zooming, elastic \\deformations.} & Dice loss & RMSprop & None & \tabincell{l}{largest connected \\component for liver}& Ensemble of three models\\ [-2pt]

\midrule
\tabincell{l}{\crule[888700]{0.15cm}{0.15cm} \textbf{K. Roth}; \\ T. Konopczynski, \\ J. Hesser} & \tabincell{l}{2D and 3D U-Net, however with \\ an iterative Mask Mining process \\ similar to model boosting}. & \tabincell{l}{flipping, rotation, \\and zooming} & \tabincell{l}{Mixture of smooth Dice loss\\ and weighted cross-entropy}& Adam & \tabincell{l}{Clipping HU values \\ to $[$-100, 600$]$} & None& None\\[-2pt]

\midrule
\crule[ff0000]{0.15cm}{0.15cm} \textbf{X. Han};  & \tabincell{l}{Residual U-Net with 2.5D \\ input (a stack of 5 slices)} & cropping, flipping &  weighted cross-entropy & \tabincell{l}{SGD with \\ momentum} & \tabincell{l}{Value-clipping to  \\ $[$-200, 200$]$}& \tabincell{l}{Connected components with \\maximum probability below \\0.8 were removed} & None\\[-2pt]

\midrule
\tabincell{l}{\crule[ff00ff]{0.15cm}{0.15cm} \textbf{J. Lipkova}; \\ M. Rempfler, \\J. Lowengrub }  & \tabincell{l}{ U-Net for liver segmentation\\ and Cahn-Hilliard Phase field \\ separation for lesions} & None & \tabincell{l}{Liver:cross-entropy; \\ Tumor:
Energy function} &  SGD & None & None & None \\ [-2pt]

\midrule
\tabincell{l}{\crule[f3f00f]{0.15cm}{0.15cm}~\textbf{L. Bi}; \\ Jinman Kim } & \tabincell{l}{Cascaded ResNet based on a \\ pre-trained FCN on 2D axial slices.} & \tabincell{l}{Random scaling, \\crops and flips} & cross-entropy & SGD & \tabincell{l}{Value-clipping to \\ $[$-160, 240$]$} &  \tabincell{l}{Morphological filter \\ to fill the holes}& \tabincell{l}{Multi-scale ensembling by \\averaging the outputs from. \\ different inputs sizes.}\\ [-2pt]

\midrule
\tabincell{l}{\crule[0261ff]{0.15cm}{0.15cm} \textbf{M. Piraud}; \\ A. Sekuboyina, \\ B. Menze } & \tabincell{l}{U-Net with a double \\ sigmoid activation} & None  & weighted cross-entropy & Adam & \tabincell{l}{Value-clipping to \\ $[$-100, 400$]$; \\intensity normalization} & None & None\\[-2pt]

\midrule
\tabincell{l}{\crule[b000b0]{0.15cm}{0.15cm} \textbf{J. Ma}; \\
Y. Li, Y. Wu, \\
M. Zhang, X. Yang} &  \tabincell{l}{Label propogation \\(Interactive method)} &  None & None & None & \tabincell{l}{Value-clipping to \\ $[$-100, 400$]$; \\intensity normalization} & None & None\\[-2pt]

\midrule
\tabincell{l}{\crule[ff1f0f]{0.15cm}{0.15cm}  \textbf{L. Zhang}; \\
S. C. YU} & \tabincell{l}{Context-aware PolyUNet with \\zooming out/in and two-stage \\strategy \citep{zhang2021context}} & Weighted cross-entropy & SGD & \tabincell{l}{Value-clipping to \\ $[$-200, 300$]$} & largest connected component & None\\[-2pt]

\bottomrule
\end{tabular}
\label{tab:method_2}
\end{table}
\end{landscape}



\subsection{Results of inidividual challenges}

\color{black}

At ISBI 2017 and MICCAI 2017, the LiTS challenges received 61 valid submissions and 32 contributing short papers as part of the two workshops. 
At MICCAI 2018, LiTS was held as part of the Medical Segmentation Decathlon and received 18 submissions (one of them was excluded from the analysis as requested by the participating team). 
In this work, the segmentation results were evaluated based on the same metrics described to ensure the comparability between the three events. 
For ISBI 2017, no liver segmentation task was evaluated. 
The results of the tumor segmentation task were shown for all the events, i.e., ISBI 2017, MICCAI 2017, and MICCAI 2018.

\subsubsection{Liver segmentation}

\paragraph{Overview} The results of the liver segmentation task showed high Dice scores; most teams achieved more than 0.930. It indicated that solely using the Dice score could not distinguish a clear winner. When we compared the progress with the two LiTS benchmarks, the results of MICCAI 2017 were slightly better than the MICCAI-MSD 2018 in terms of ASD (1.104 \emph{vs.} 1.342). It might be because the algorithms for MICCAI-MSD were optimized considering their generalizability on different organs and imaging modalities. In contrast, the methods for MICCAI 2017 were specifically optimized for liver and CT imaging. The ranking result is shown in Table~\ref{tab:lits_liver_segmentation}.

\paragraph{LiTS--MICCAI 2017}
The evaluation of the liver segmentation task relied on the three metrics explained in the previous chapter, with the Dice score per case acting as the primary metric used for the final ranking. Almost all methods except the last three achieved Dice per case values above 0.920, with the best one scoring 0.963. Ranking positions remain relatively stable when ordering submissions according to the other surface distance metric. Most methods changed by a few spots, and the top four methods were only interchanging positions among themselves. The position variation was more significant than the Dice score when using the surface distance metric ASD for the ranking, with some methods moving up to 4 positions. 
However, on average, the top-2 performing Dice per case methods still achieved the lowest surface distance values, with the winning method retaining the top spot in two rankings. 

\paragraph{LiTS--MICCAI--MSD 2018}
The performance of Decathlon methods in liver segmentation showed similar results compared to MICCAI 2017. In both challenges, one could observe that the difference in Dice scores between top-performing methods was insignificant mainly because the liver is a large organ. The comparison of ASD between MICCAI 2018 and MICCAI 2017 confirmed that state-of-the-art methods could automatically segment the liver with similar performance to manual expert annotation for most cases. However, the methods exclusively trained for liver segmentation in MICCAI 2017 showed better segmentations under challenging cases (1.104 vs. 1.342).

\subsubsection{Liver tumor segmentation and detection}

\paragraph{Overview} While automated liver segmentation methods showed promising results (comparable to expert annotation), the liver tumor segmentation task remained room for improvement. 
To illustrate the difficulty of detecting small lesions, we grouped the lesions into three categories with a clinical standard: a) small lesions less than 10 mm in diameter, b) medium lesions between 10 mm and 20 mm in diameter, and c) large lesion bigger than 20 mm in diameter. The ranking result is shown in Table~\ref{tab:lits_lesion}.

\begin{figure}[!t]
\centering
\includegraphics[width=0.95\textwidth]{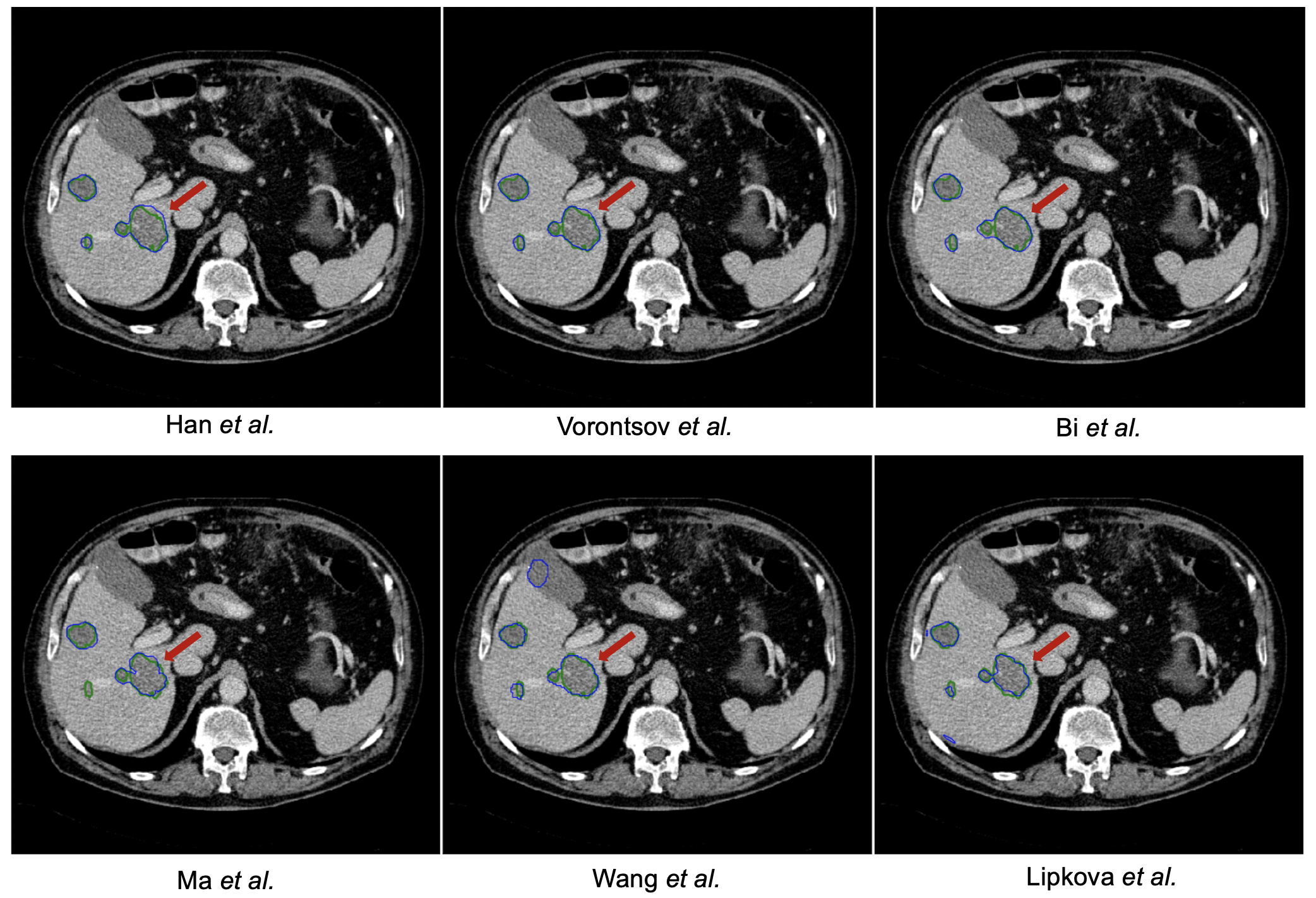}
\caption{Tumor segmentation results of the ISBI--LiTS 2017 challenge. The reference annotation is marked with green contour, while the prediction is with blue contour.  {One could observe that the boundary of liver lesion is rather. ambiguous.}}
\label{fig:imagesISBI}
\end{figure}

\paragraph{LiTS--ISBI 2017}
The highest Dice scores for liver tumor segmentation were in the middle 0.60s range, with the winner team achieving a score of 0.674 followed by 0.652 and 0.645 for the second and third places, respectively. However, there were no statistically significant differences between the top three teams in all three metrics. The final ranking changed to some degree when considering the ASD metric. For example, Bi~\emph{et~al.} obtained the best ASD score but retained its order with the best methods overall. In lesion detection, we found that detecting small lesions was very challenging in which top-performing teams achieved only around 0.10 in F1 score. Figure \ref{fig:imagesISBI} shows some sample results of the top-performing methods.

\begin{figure}[!t]
\centering
\includegraphics[width=0.95\textwidth]{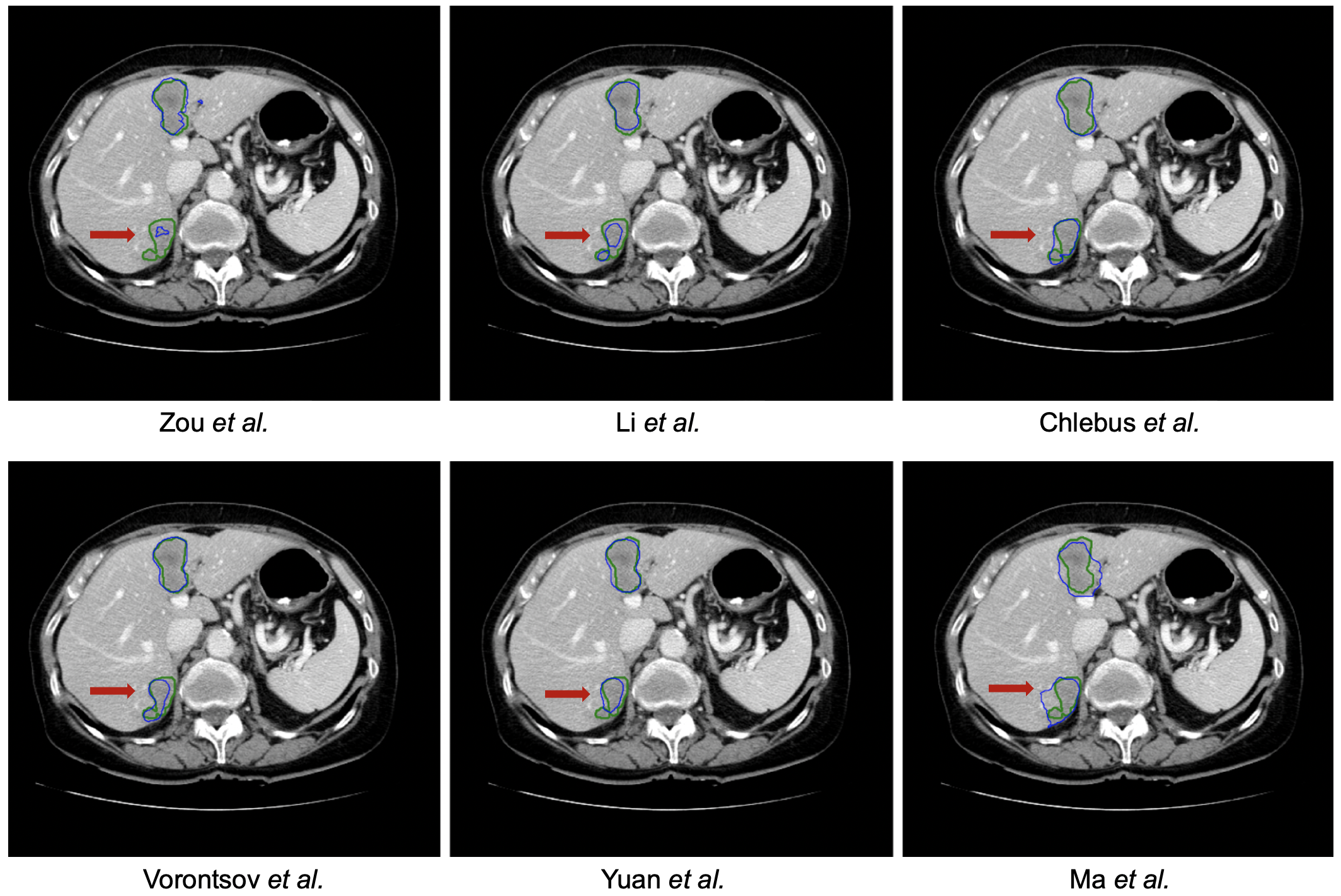}
\caption{Tumor segmentation results of the MICCAI--LiTS 2017 challenge. The reference annotation is marked with green contour, while the prediction is with blue contour. One could observe that it is highly challenging to segment the liver lesion with poor contrast.}
\label{fig:imagesMICCAI}
\end{figure}

\paragraph{LiTS--MICCAI 2017}
The best tumor segmentation Dice scores improved significantly compared to ISBI, with MICCAI's highest average Dice (per case) of 0.702 compared to 0.674 in ISBI on the same test set. However, the ASD metric did not improve (1.189 \emph{vs.} 1.118) on the best top-performing method. In addition, there were no statistically significant differences between the top three teams in all three metrics.
There was an overall positive correlation of ranking positions with submissions that performed well at the liver segmentation task concerning the liver tumor segmentation task. A weak positive correlation between the Dice ranking and the surface distance metrics could still be observed, although a considerable portion of methods changes positions by more than a few spots. The detection performance in MICCAI 2017 showed improvement over ISBI 2017 in lesion recall (0.479 \emph{vs.} 0.458 for the best team). Notably, the best-performing team (\crule[baba00]{0.20cm}{0.20cm} J. Zou \emph{et al.}) achieved a very low precision of 0.148, which indicated that the method generates many false positives. Figure \ref{fig:imagesMICCAI} shows some sample results of the top-performing methods. 

\paragraph{LiTS--MICCAI--MSD 2018}
The LiTS evaluation of MICCAI 2018 was integrated into MSD and attracted much attention, receiving 18 valid submissions. Methods were ranked according to two metrics: Dice score and ASD (in liver and liver tumor segmentation tasks). Compared to MICCAI 2017 and ISBI 2016, the two top-performing teams significantly improved the Dice scores (0.739 and 0.721 \emph{vs.} 0.702) and ASD (0.903 and 0.896 \emph{vs.} 1.189). However, there were no statistically significant differences between the top two teams in all three metrics. The first place (F. Isensee \emph{et al.}) statistically significant (p-value $<$ 0.001) outperformed the third place (S. Chen \emph{et al.}) considering Dice score. More importantly, the same team won the two tasks using a self-adapted 3D deep learning solution, indicating a step forward in the development of segmentation methodology. The detection performance in MICCAI 2018 showed improvement over MICCAI 2017 in lesion recall (0.554 \emph{vs.} 0.479 for the best-performing teams).

\begin{figure}[!h]
\centering
\includegraphics[width=0.95\textwidth]{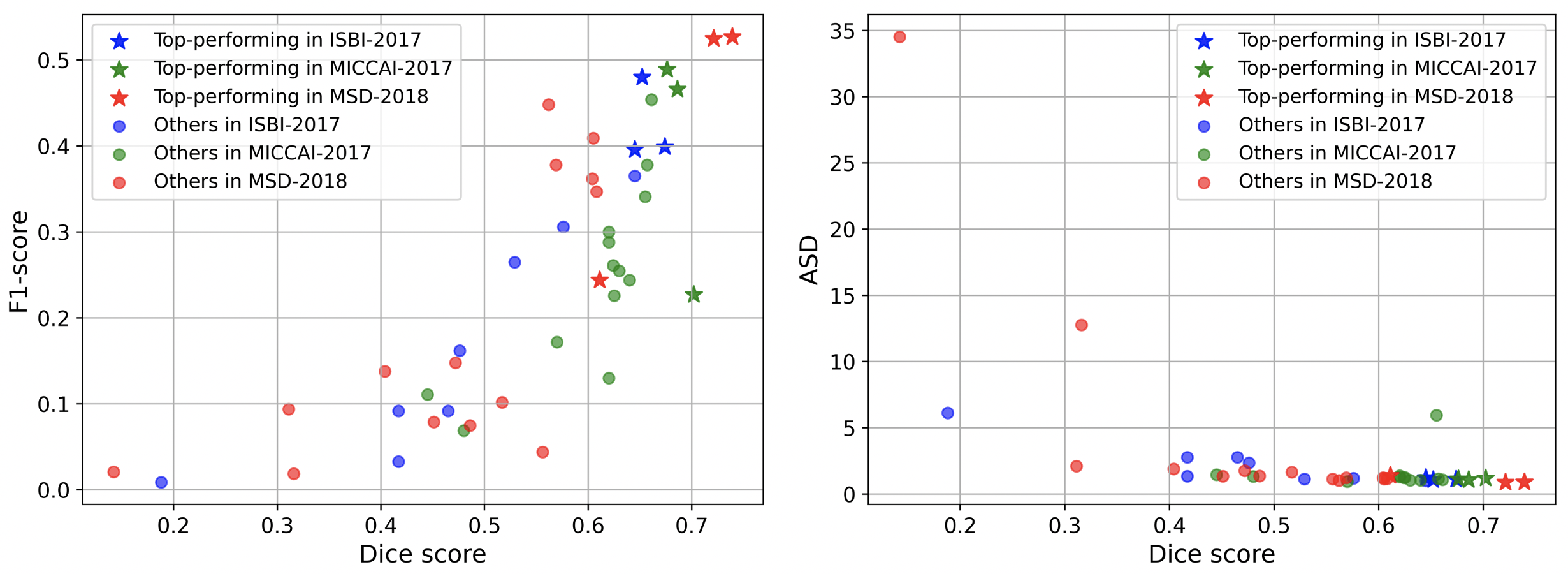}
\caption{Scatter plots of methods' performances considering: a) both segmentation and detection, b) both distance- and overlap-based metrics for three challenge events. We observe that not all the top-performing methods in three LiTS challenges achieved good scores on tumor detection. The behavior of distance- and overlap-based metrics is similar.}
\label{fig:scatter_plot}
\end{figure}

From the scatter plots shown in Figure \ref{fig:scatter_plot}, we observed that not all the top-performing methods in three LiTS challenges achieved good scores on tumor detection. The behavior of distance- and overlap-based metrics was similar. The detection metrics with clinical relevance could prevent the segmentation model from tending to segment large lesions. Thus it should be considered when ranking participating teams and performing the comprehensive assessment. 



\begin{table*}[!htbp]
\tiny
\centering
\begin{tabular}{l | l | l | l | l | l }
\specialrule{.1em}{0em}{-.1em}
 & \textbf{Ranking}~&~\textbf{Ref. Name} &\textbf{Dice} &\textbf{ASD}  & \textbf{{Re-ranking*}} \\ [0.5ex] 
\specialrule{.05em}{-0.1em}{0em}
\parbox[t]{1mm}{\multirow{17}{*}{\rotatebox[origin=c]{90}{\textsc{LiTS--MICCAI} 2017}}} 
& 1  & Y. Yuan \emph{et al.}  & \cellcolor{blue!10}0.963 (1)  & \cellcolor{blue!10}1.104 (1) & 2~~({\textbf{1}}) \\
& 2 & A. Ben-Cohen \emph{et al.}  & \cellcolor{blue!10}0.962 (2) & \cellcolor{blue!10}1.130 (2)  & 4~~(\textbf{2})\\
& 3  & J. Zou \emph{et al.}    & \cellcolor{blue!10}0.961 (3) & 1.268 (4)  & 7~~({\textbf{3}})\\
& 4   & X. Li \emph{et al.}   & \cellcolor{blue!10}0.961 (3)  & 1.692 (8) & 10 (4) \\
& 5  & L. Zhang \emph{et al.}   & 0.960 (4) & 1.510 (7)  & 11~~({\textbf{5}})\\
& 6  & G. Chlebus \emph{et al.}   & 0.960 (4)    & \cellcolor{blue!10}1.150 (3)  & 7~~({\textbf{3}})\\

& 7  &J. Wu \emph{et al.}   & 0.959 (5)   & 1.311 (5)  & 10 (4)\\
& 8  & C. Wang \emph{et al.} & 0.958 (6)     & 1.367 (6)  & 12 (6)\\
& 9 & E. Vorontsov \emph{et al.}  & 0.951 (7)  & 1.785 (9) & 16 (7)\\
& 10 & K. Kaluva \emph{et al.}      & 0.950 (8)  & 1.880 (10) & 18 (8)\\
& 11 & K. Roth \emph{et al.}   & 0.946 (9) & 1.890 (11)  & 20 (9)\\
& 12 &X. Han \emph{et al.}   & 0.943 (10)  & 2.890 (12)  & 22 (10)\\
& 13   & J. Lipkova \emph{et al.}      & 0.938 (11) & 3.540 (13) &  24 (11)\\
& 14   & L. Bi \emph{et al.}  & 0.934 (12) & 258.598 (15)  & 26 (12)\\
& 15 & M. Piraud \emph{et al.}     & 0.767 (13)  & 37.450 (14)  & 26 (12)\\
& 16 & J. Ma \emph{et al.}    & 0.041 (14)  & 8231.318 (15) & 29 (13)\\ 

\specialrule{.05em}{-0.1em}{0em}
\parbox[t]{1mm}{\multirow{17}{*}{\rotatebox[origin=c]{90}{\textsc{LiTS-MICCAI}~2018}}} 

&1 &F. Isensee \emph{et al.}      &	\cellcolor{blue!10}0.962 (1)    &	2.565 (9)  & 10 (5)\\
&2 & Z. Xu \emph{et al.}    & \cellcolor{blue!10}0.959 (2)   &	\cellcolor{blue!10}1.342 (1)	& 3~~({\textbf{1}}) \\
&3&	D. Xu \emph{et al.} & \cellcolor{blue!10}0.959 (2)  &	1.722 (5)   & 7~~({\textbf{3}}) \\
&4  &	B. Park \emph{et al.}& 0.955 (3)   &	1.651 (4)  & 7~~({\textbf{3}})\\
&5	    &S. Chen \emph{et al.} & 0.954 (4)   &\cellcolor{blue!10}1.386 (2)  & 6~~({\textbf{2}})\\
&6	 &	R. Chen \emph{et al.} & 0.954 (5) &	\cellcolor{blue!10}1.435 (3)  & 8~~(4)\\
&7	 &	M. Perslev \emph{et al.} & 0.953 (6)   &	2.360 (8)  & 14 (6)\\
&8	    &	I. Kim \emph{et al.}   & 0.948 (7)    &	2.942 (12)  & 19 (8)\\
&9	  &	O. Kodym \emph{et al.}	   & 0.942 (8)    &	2.710 (11) & 19 (8)\\
&10	 &F. Jia \emph{et al.}   & 0.942 (9)   &	2.198 (6)  & 15~(7)\\
&11	    & S. Kim \emph{et al.} & 0.934 (10)    &	6.937 (14) & 24 (10)\\
&12	  &	W. Bae \emph{et al.} & 0.934 (11)  &	2.615 (10) & 21 (9)\\
&13	  &	Y. Wang \emph{et al.} 	  & 0.926 (12)   &	2.313 (7)  & 19 (8)\\
&14	&I. Sarasua \emph{et al.} & 0.924 (13) &	6.273 (13) 	&26 (11)\\
&15	& O. Rippel \emph{et al.}   & 0.904 (14)       &	16.163 (16) & 30 (12)\\
&16	  &	R. Rezaeifar \emph{et al.}   & 0.864 (15)    &	9.358 (15)  & 30 (12)\\
&17	    &	J. Ma \emph{et al.}	 & 0.706 (16)   &	159.314 (17) & 33 (13)\\
\bottomrule
\end{tabular}%

\caption{
Liver segmentation submissions of LiTS--MICCAI 2017 and LiTS--MICCAI--MSD 2018 ranked by Dice score and ASD. Top-performing teams in liver segmentation tasks are highlighted with blue in each metric. {A ranking score follows each metric value in the bracket. {\emph{Re-ranking*} denotes a post-challenge ranking considering two metrics by averaging the ranking scores. Notably only Dice and RVD are considered as the volume difference of the liver is not of interest, as opposed to the liver tumor.}}}

\label{tab:lits_liver_segmentation}
\end{table*}

\begin{table*}[!htbp]
\tiny
\centering
\setlength\tabcolsep{2pt} 
\begin{tabular}{l | l | l | l  | l | l | l || l |l | l | l | l }
\specialrule{.1em}{0em}{-.1em}
 & \textbf{Ranking}~&~\textbf{Ref. Name} &\textbf{Dice} &\textbf{ASD}  &\textbf{RVD} & \textbf{ {Re-ranking*}} &\textbf{Precison}  &\textbf{Recall} & \textbf{F1}$_{small}$ & \textbf{F1}$_{medium}$ & \textbf{F1}$_{large}$\\ [0.5ex] 
\specialrule{.05em}{-0.1em}{0em}
\parbox[t]{1mm}{\multirow{14}{*}{\rotatebox[origin=c]{90}{\textsc{LiTS--ISBI} 2017}}}

& 1  & X. Han \emph{et al.} &\cellcolor{blue!10}0.674 (1) & \cellcolor{blue!10}1.118 (3) &   -0.103 (7)  & 11 ({\textbf{3}}) & 0.354 (4) & \cellcolor{red!10}0.458 (1) & \cellcolor{red!10}0.103 (3) & \cellcolor{red!10}0.450 (1) & \cellcolor{red!10}0.879 (1)  \\
& 2 & G. Chlebus \emph{et al.}   &\cellcolor{blue!10}0.652 (2)& \cellcolor{blue!10}1.076 (2)  & \cellcolor{blue!10}-0.025 (2) & 6~~({\textbf{2}}) & \cellcolor{red!10}0.385 (3) & 0.406 (4) & \cellcolor{red!10}0.116 (1) & \cellcolor{red!10}0.421 (3) & \cellcolor{red!10}0.867 (3)  \\
& 3 & E. Vorontsov \emph{et al.}   &\cellcolor{blue!10}0.645 (3)  & 1.225 (6)  & -0.124 (8) & 17 (6) & \cellcolor{red!10}0.529 (2) & \cellcolor{red!10}0.439 (2) & \cellcolor{red!10}0.109 (2) & 0.369 (4) & 0.850 (4) \\
& 4 &L. Bi \emph{et al.}  &\cellcolor{blue!10}0.645 (3) & \cellcolor{blue!10}1.006 (1)  & \cellcolor{blue!10}0.016 (1) & 5~~({\textbf{1}})  & 0.316 (5) & \cellcolor{red!10}0.431 (3) & 0.057 (5) & \cellcolor{red!10}0.441 (2) &  \cellcolor{red!10}0.876 (2) \\ 
& 5 & C. Wang \emph{et al.}   &0.576 (4)  & 1.187 (5) & -0.073 (5) & 14 (5)  & 0.273 (6) & 0.346 (7) & 0.038 (7) & 0.323 (5) & 0.806 (7) \\
& 6  & P. Christ \emph{et al.}  &0.529 (5) & 1.130 (4) & \cellcolor{blue!10}0.031 (3) & 12 (4) & \cellcolor{red!10}0.552 (1) & 0.383 (5) & 0.075 (4) & 0.221 (7) & 0.837 (6) \\ 
& 7 & J. Lipkova \emph{et al.}    &0.476 (6)  & 2.366 (8)  &-0.088 (6) & 20 (7) & 0.105 (8)  & 0.357 (6) & 0.054 (6) & 0.250 (6) & 0.843 (5) \\
&8 & J. Ma \emph{et al.}   &0.465 (7) & 2.778 (10)  & 0.045 (4) &21 (8)  & 0.107 (7)   & 0.080 (10) & 0.000 (10) & 0.033 (10) & 0.361 (10) \\
&9 & T. Konopczynski \emph{et al.}    &0.417 (8) & 1.342 (7) &-0.150 (9) & 24 (9) & 0.057 (9) & 0.189 (8) & 0.014 (9) & 0.106 (8)  & 0.628 (8)\\
&10  & M. Bellver \emph{et al.}   &0.411  (9)  & 2.776 (9)  &-0.263 (11) & 29 (10) & 0.028 (10) &  0.172 (9) & 0.031 (8) & 0.175 (9) & 0.512 (9)\\
&11 & J. Qi \emph{et al.}  &0.188 (10)  & 6.118 (11) &-0.229 (10) & 31 (11)& 0.008 (11) & 0.009 (11) & 0.000 (10) & 0.000 (11) & 0.041 (11) \\ 


\specialrule{.05em}{-0.1em}{0em}
\parbox[t]{1mm}{\multirow{16}{*}{\rotatebox[origin=c]{90}{\textsc{LiTS--MICCAI} 2017}}} 

&1 & J. Zou \emph{et al.}  &\cellcolor{blue!10}0.702 (1)   & 1.189 (8) & 5.921 (10) & 14 ({\textbf{2}}) & 0.148 (13) & 0.479 \cellcolor{red!10}(2) & \cellcolor{red!10}0.163 (2) & 0.446 (4) & 0.876 (5)\\
&2 & X. Li \emph{et al.} & \cellcolor{blue!10}0.686 (2) & 1.073 (4) & 5.164 (9) & 16 (4) & \cellcolor{red!10}0.426 (3)& \cellcolor{red!10}0.515 (1) & 0.150 (4) & \cellcolor{red!10}0.544 (1) & \cellcolor{red!10}0.907 (3) \\
&3  & G. Chlebus \emph{et al.}    & \cellcolor{blue!10}0.676 (3)    & 1.143 (6) &0.464 (7) & 16 (4) & \cellcolor{red!10}0.519 (1) & \cellcolor{red!10}0.463 (3) & 0.129 (6)& \cellcolor{red!10}0.494 (2)  & 0.836 (10)\\
&4 & E. Vorontsov \emph{et al.}   & 0.661 (4)   & 1.075 (5)  & 12.124 (15) & 24 (6) & \cellcolor{red!10}0.454 (2) & 0.455 (6) & 0.142 (5) & 0.439 (6) & 0.877 (4)\\
&5  & Y. Yuan \emph{et al.}     & 0.657 (5)   & 1.151 (7) &0.288 (6) & 18 (5) & 0.321 (5) & 0.460 (5) & 0.112 (8) & \cellcolor{red!10}0.471 (3) & 0.870 (8)\\
&6  & J. Ma \emph{et al.}       & 0.655 (6) &5.949 (12) & 5.949 (11) & 29 (9)  & 0.409 (4) & 0.293 (14) & 0.024 (13) & 0.200 (13)& 0.770 (13)\\
&7  & K. Kaluva \emph{et al.}     & 0.640 (7)     & \cellcolor{blue!10}1.040 (2) &0.190 (4)  & 13 ({\textbf{1}}) & 0.165 (10) & 0.463 (4) & 0.112 (7) &  0.421 (8)& \cellcolor{red!10}0.910 (1)\\
&8  & X. Han~\emph{et al.}     & 0.630 (8) & \cellcolor{blue!10}1.050 (3) &\cellcolor{blue!10}0.170 (3) & 14 ({\textbf{2}}) & 0.160 (11)  & 0.330 (11) & 0.129 (6) & 0.411 (9) & \cellcolor{red!10}0.908 (2)\\
&9 & C. Wang~\emph{et al.}  & 0.625 (9) & 1.260 (10)  &8.300 (13) & 32 (10) & 0.156 (12) & 0.408 (8) & 0.081 (10) & 0.423 (7)& 0.832 (11)\\
&10  & J. Wu \emph{et al.} & 0.624 (10) & 1.232 (9) &4.679 (8) & 27 (7) & 0.179 (9) & 0.372 (9) & 0.093 (9) & 0.373 (11)& 0.875 (6)\\
&11 & A. Ben-Cohen \emph{et al.}  & 0.620 (11)   & 1.290 (11) &0.200 (5) & 27 (7) & 0.270 (7) & 0.290 (15) & 0.079 (11) & 0.383 (10) & 0.864 (9)\\
&12 & L. Zhang \emph{et al.} & 0.620 (12)  & 1.388 (13)  &6.420 (12) & 37 (11) & 0.239 (8) & 0.446 (7) & \cellcolor{red!10}0.152 (3) & 0.445 (5) & 0.872 (7)\\
&13      & K. Roth \emph{et al.}   & 0.570 (13)     & \cellcolor{blue!10}0.950 (1)  &\cellcolor{blue!10}0.020 (1) & 15 ({\textbf{3}}) & 0.070 (14) & 0.300 (13) &  \cellcolor{red!10}0.167 (1)&  0.411 (9)& 0.786 (12)\\
&14      & J. Lipkova \emph{et al.}    & 0.480 (14)  & 1.330 (12)  &\cellcolor{blue!10}0.060 (2)  & 28 (8)&  0.060 (16) & 0.190 (16) & 0.014 (14) & 0.206 (12) & 0.755 (13)\\
&15      & M. Piraud \emph{et al.}    & 0.445 (15) & 1.464 (14) &10.121 (14) & 43 (12) & 0.068 (15) & 0.325 (12) & 0.038 (12) & 0.196 (14) & 0.738 (15)\\

\specialrule{.05em}{-0.1em}{0em}
\parbox[t]{1mm}{\multirow{19}{*}{\rotatebox[origin=c]{90}{\textsc{LiTS--MICCAI} 2018}}} 

&1	   &F. Isensee \emph{et al.}  &	\cellcolor{blue!10}0.739 (1)   &	\cellcolor{blue!10}0.903 (2)  & -0.074 (10) & 13 ({\textbf{2}}) & 0.502 (4) & \cellcolor{red!10}0.554 (1) & \cellcolor{red!10}0.239 (1) & \cellcolor{red!10}0.564 (1) & \cellcolor{red!10}0.915 (2)\\
&2	    &	D. Xu \emph{et al.}  &	\cellcolor{blue!10}0.721 (2)   &	\cellcolor{blue!10}0.896 (1) & \cellcolor{blue!10}-0.002 (1) & 4~~({\textbf{1}})&  \cellcolor{red!10}0.549 (2) & \cellcolor{red!10}0.503 (2) & \cellcolor{red!10}0.149 (3) & \cellcolor{red!10}0.475 (2) & \cellcolor{red!10}0.937 (1)\\
&3	    &S. Chen \emph{et al.}	  &	\cellcolor{blue!10}0.611 (3)   & 1.397 (11)  &-0.113 (12)& 26 (6) &0.182 (9) & 0.368 (4) & 0.035 (9) & 0.239 (12) & \cellcolor{red!10}0.859 (3) \\
&4	    &B. Park \emph{et al.}  & 0.608 (4)    &	1.157 (6) 	& -0.067 (8) & 18 (5) & 0.343 (5) & 0.350 (7) & 0.044 (8) & 0.267 (7) & 0.845 (4)\\
&5	    &	O. Kodym \emph{et al.}   &	0.605 (5)   &	1.134 (4)  &-0.048 (6) & 16 ({\textbf{3}}) &  \cellcolor{red!10}0.523 (3) & 0.336 (8) & 0.063 (7) & 0.243 (10) & 0.819 (6)\\
&6	    &	Z. Xu \emph{et al.}  &	0.604 (6)  &	1.240 (8)  & -0.025 (4)& 18 (5) & 0.396 (6) & 0.334 (9) & 0.015 (10)& 0.243 (11) & 0.837 (5)\\
&7	    &R. Chen \emph{et al.}  &	0.569 (7)  &	1.238 (7)  & -0.188 (14) & 28 (8) &  0.339 (7) & \cellcolor{red!10}0.427 (3)  & \cellcolor{red!10}0.207 (2)& \cellcolor{red!10}0.366 (3)  & 0.804 (8)\\
&8	    &		I. Kim \emph{et al.}    &	0.562 (8)   &	\cellcolor{blue!10}1.029 (3) &\cellcolor{blue!10}0.012 (2)  & 13 (2) & \cellcolor{red!10}0.594 (1) & 0.360  (5) & 0.092 (4) & 0.328 (5) & 0.781 (9)\\
&9	    & M. Perslev \emph{et al.}	  &	0.556 (9)  &	1.134 (5) & \cellcolor{blue!10}0.020 (3)& 17 (4)&  0.024 (17) & 0.330 (10) & 0.034 (10) & 0.251 (8)  & 0.811 (7)\\
&10	    &W. Bae \emph{et al.}	    &	0.517 (10)  &	1.650 (12) &-0.039 (5) & 27 (7) &  0.061 (14) & 0.308 (11) & 0.078 (6) & 0.244 (9) & 0.742 (11)\\
&11	    &I. Sarasua \emph{et al.}  &	0.486 (11)  &	1.374 (10)  &-0.084 (11)  & 32 (10) & 0.043 (16) & 0.298 (12) & 0.045 (8) & 0.294 (6)  & 0.678 (12)\\
&12	    &R. Rezaeifar \emph{et al.}    &	0.472 (12)    &	1.776 (13)  &-0.258 (15) & 40 (12) & 0.112 (11) & 0.216 (13) & 0.005 (12) & 0.081 (14)  & 0.650 (13)\\
&13	    &	O. Rippel \emph{et al.}  &	0.451 (13)  &	1.345 (9)&-0.068 (9) & 31 (9) & 0.044 (15) & 0.356 (6) & 0.083 (5)  & 0.353 (4) & 0.771 (10)\\
&14	    &	S. Kim \emph{et al.}  &0.404 (14)   &	1.891 (14)  &0.151 (13) & 41 (13)& 0.116 (10) & 0.170 (14) & 0.005 (12) &  0.091 (13)&  0.589 (14)\\
&15	    &F. Jia \emph{et al.} 	   &	0.316 (15)   &	12.762 (16) &-0.620 (16) &47 (14)  & 0.069 (12) & 0.011 (17)& 0.015 (10) & 0.000 (16) & 0.024 (17)\\
&16	    &Y. Wang \emph{et al.} &	0.311 (16)  &	2.105 (15)  &0.054 (7) &38 (11) & 0.154 (8) & 0.068 (15) & 0.000 (13) & 0.005 (15) & 0.336 (15)\\
&17	    &	J. Ma \emph{et al.}	  &	0.142 (17) &	34.527 (17) & 0.685 (17)& 51 (15) & -0.066 (13) & 0.013 (16) & 0.000 (13) & 0.000 (16) & 0.049 (16)\\

\bottomrule
\end{tabular}

\caption{Liver tumor segmentation results of three challenges ranked by segmentation metrics (i.e., Dice, ASD and RVD) and detection metrics (i.e., precision, recall and separated F1 scores with three different sizes of lesion). {Each metric value is followed by a ranking score in the bracket.} Top performing teams in tumor segmentation and tumor detection are highlighted with blue and pink colors in each metric, respectively.  {\emph{Re-ranking*} denotes a post-challenge ranking considering three metrics {by averaging the ranking scores}.}}
\label{tab:lits_lesion}
\end{table*}

\subsection{ {Meta Analysis}}
\begin{figure}[!t]
\centering
\includegraphics[width=0.50\textwidth]{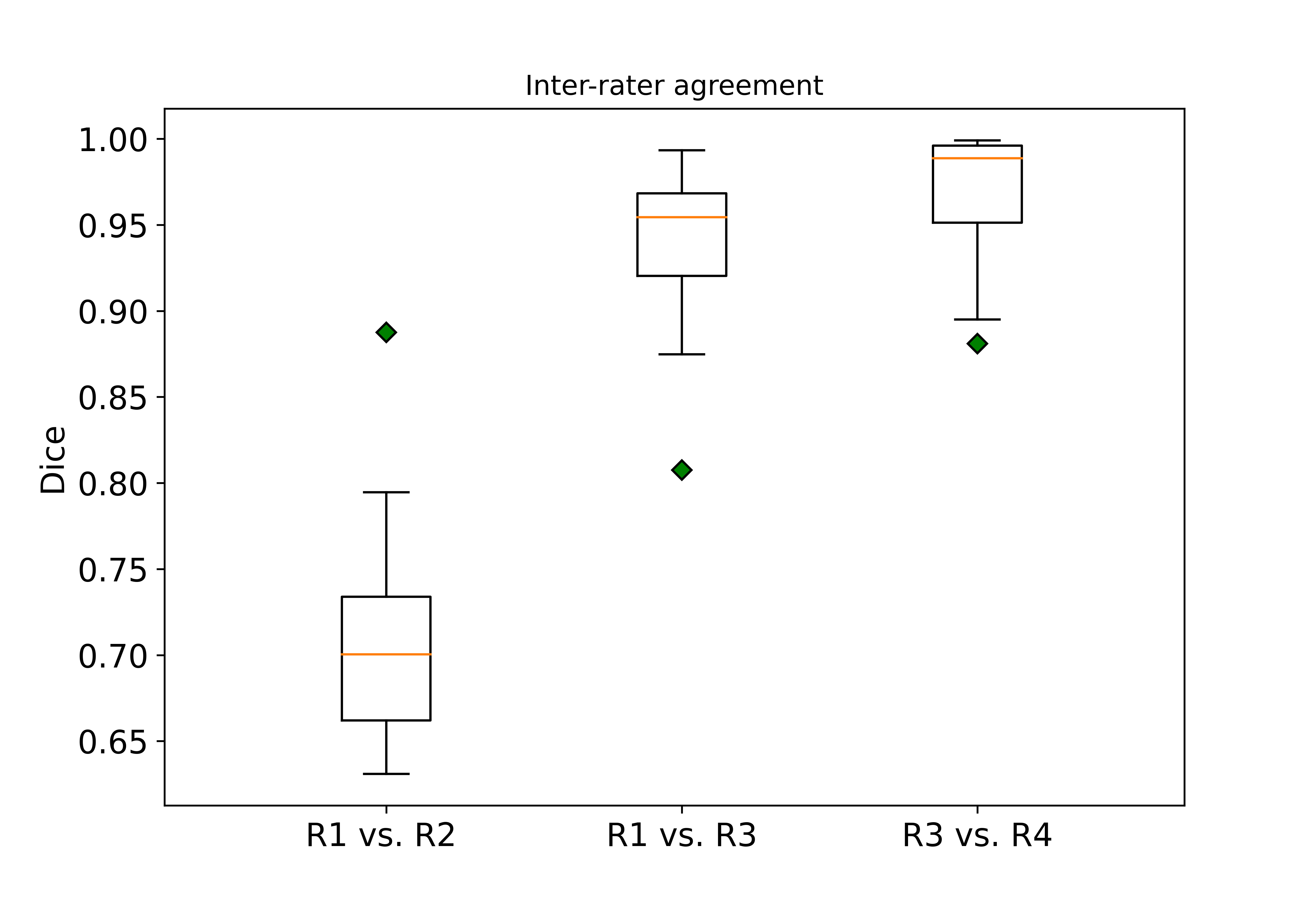}
\caption{{Inter-rater agreement between the existing annotation and new annotation sets. R1 represented the rater for the existing consensus annotation of the LiTS dataset. R2 re-annotated 15 CT scans from scratch. R3 and R4 are board-certified radiologists who checked and corrected the annotations. Specifically, one board-certified radiologist (R3) reviewed and corrected existing annotations. R4 re-evaluated R3's final annotations and corrected them. The inter-rater agreement was calculated by the Dice score per case between the pairs of two raters.}}
\label{fig:inter_rater}
\end{figure}

 {In this section, we focus on liver tumor segmentation and analyze the inter-rater variability and method development during the last six years.}
 {\subsubsection{Inter-rater agreement} To better interpret the algorithmic variability and performance, we recruited another radiologist (Z. Z.) with $>$3 years of experience in oncologic imaging to re-annotate 15 3D CT scans, and two board-certified radiologists (J. K. and B. W.) to re-evaluate the original annotations. In Figure \ref{fig:inter_rater}, R2 re-annotated 15 CT scans from scratch. R3 and R4 are board-certified radiologists who checked and corrected the annotations. Specifically, one board-certified radiologist (R3) reviewed and corrected existing annotations. R4 re-evaluated R3's final annotations and corrected them. The inter-rater agreement was calculated by the Dice score per case between the pairs of two raters. We observed high inter-rater variability (median Dice of 70.2\%)between the new annotation (R2) and the existing consensus annotation. We observed very high agreement (median Dice of 95.2\%) between the board-certified radiologist and the existing annotations. Considering that the segmentation models were solely optimized on R1 and the best model achieved 82.5\% on the leader-board (last access: 04.04.2022), we argue that there is still room for improvement. }

\begin{figure}[!t]
\centering
\includegraphics[width=0.65\textwidth]{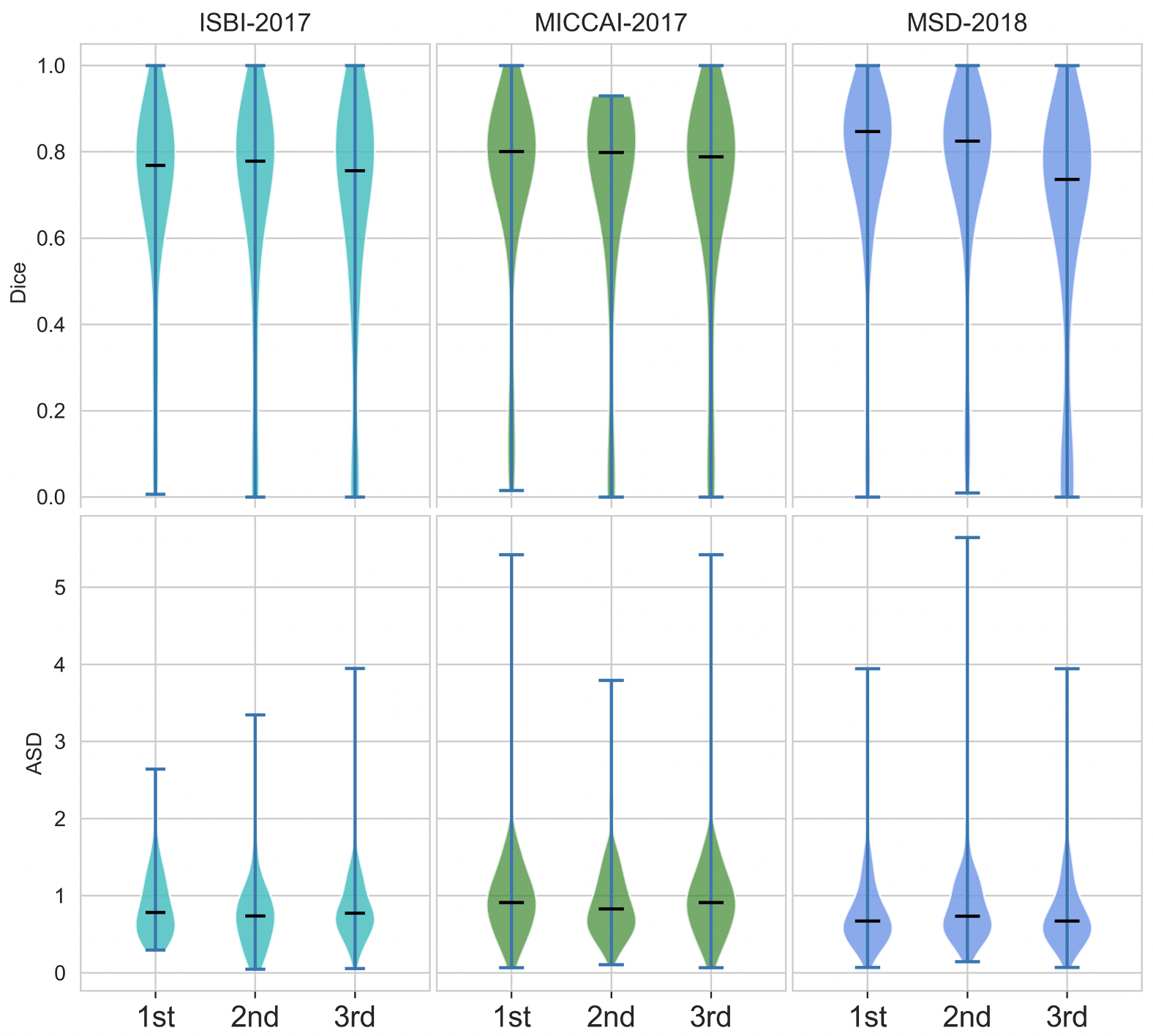}
\caption{Dice and ASD scores of three top-performing teams over the three events.}
\label{fig:violin_plot_top3}
\end{figure}

 {\subsubsection{Performance improvement}}
 {\paragraph{Top-performing teams over three events} First, we plotted the scores of Dice and ASD for three top-performing teams over the three events, as shown in Figure~\ref{fig:violin_plot_top3}. We observed incremental improvement (e.g., the median scores) over the three events. We further performed Wilcoxon signed-rank tests on the best teams (ranked by mean Dice) between each pair of two events. We observed that MSD'18 shows significant improvement against ISBI'17 on both metrics (see Table \ref{tab:statistics_improvement}). For MSD'18, the submitted algorithms architectures were the same for all sub-tasks. They were trained individually for sub-tasks (e.g., liver, kidney, pancreas), focusing on the generalizability of segmentation models. The main advance was the advent of 3D deep learning models after 2017. Tables 4 and 5 show that most of the approaches were 2D and 2.5D based. The winner of MSD - the nn-Unet approach was a 3D UNet based, self-configured and adaptive for specific tasks. We attributed the main improvement to the 3D architectures, which was in line with other challenges and benchmark results in medical image segmentation that occurred during this time.}

\begin{figure}[!t]
\centering
\includegraphics[width=0.9\textwidth]{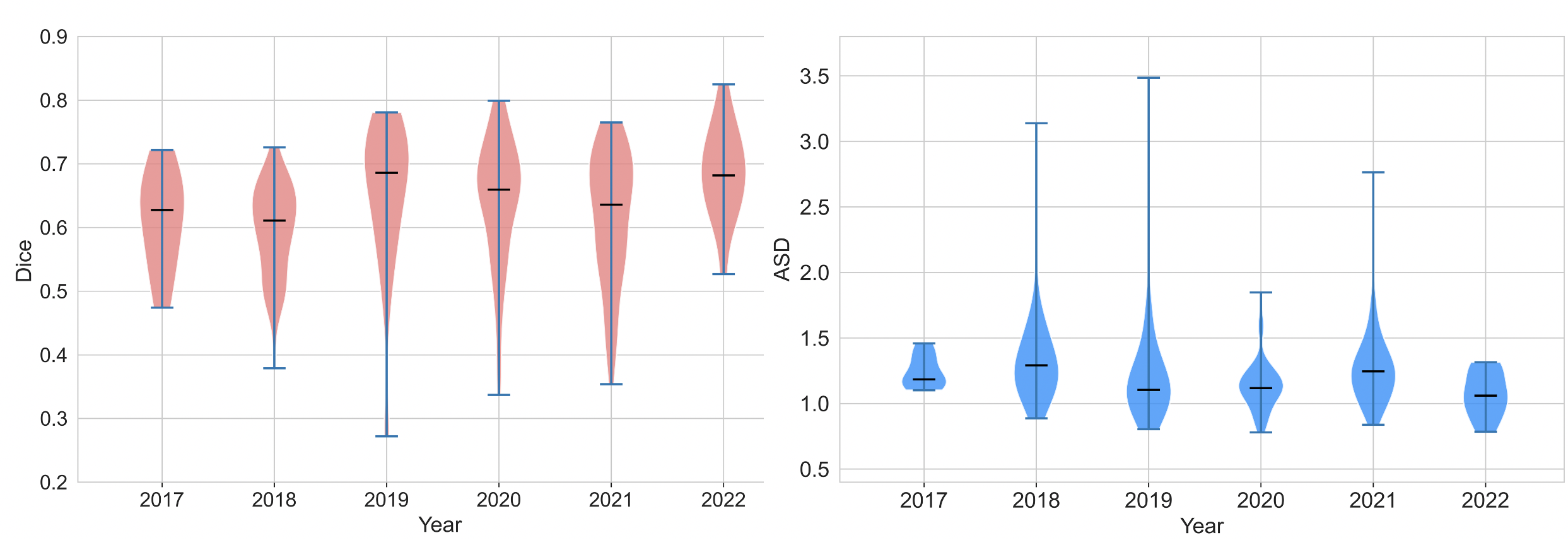}
\caption{Distribution of mean Dice and ASD scores of all submissions in the CodaLab platform from the year 2017 to the year 2022.}
\label{fig:volin_colab}
\end{figure}

 {\paragraph{CodaLab submissions in the last six years} First, we separated the submissions yearly and summarized them by individual violin plots shown in Figure \ref{fig:volin_colab}. We observed a continuous improvement over the years, with the best results obtained in 2022. We excluded the submissions that achieved Dice scores $>$10\% in the analysis.
We further performed the Mann-Whitney U test on the distributions of mean Dice and ASD scores of all teams for each year. We observed that the scores achieved in 2022 are significantly better than in 2021, indicating that the LiTS challenge remains active and contributes to methodology development.}

\begin{table*}[t]
\footnotesize
\centering
\begin{tabular}{l | c | c | c}
 \specialrule{.1em}{0em}{-.1em}
{p-value} & {MICCAI'17 vs. ISBI'17} & MSD'18 vs. MICCAI'17 & MSD'18 vs. ISBI'17 \\ [0.5ex]
\hline
 Dice & 0.236 & 0.073 & \textbf{0.015}\\
\cline{1-4}
 ASD & 0.219 & 0.144 & \textbf{0.006}\\
 \specialrule{.1em}{0em}{-.1em}
 \end{tabular}
\caption{
 {Results of Wilcoxon signed-rank tests between each pair of two events. MSD’18 significantly improved over ISBI’17 on both metrics.}}
\label{tab:statistics_improvement}
\end{table*}



\begin{table*}[t]
\scriptsize
\centering
\begin{tabular}{l | c | c | c | c | c}
 \specialrule{.1em}{0em}{-.1em}
{p-value} & {2022 vs. 2021} & 2022 vs. 2020 & 2022 vs. 2019 & 2022 vs. 2018 & 2022 vs. 2017 \\ [0.5ex]
\hline
 Dice &  \textbf{0.013} & 0.088 & 0.235 & \textbf{$<$0.001} & \textbf{0.024}
 \\
\cline{1-6}
 ASD & \textbf{0.002} & 0.158 & 0.129 & \textbf{$<$0.001} & \textbf{0.016}\\
 \specialrule{.1em}{0em}{-.1em}
 \end{tabular}
\caption{ {Results of Mann-Whitney U tests between the year of 2022 and the other years.}}
\label{tab:statistics_codalab}
\end{table*}

\subsection{ {Technique trend and recent advances}}
\label{sec:post-lits}
We have witnessed that the released LiTS dataset contributes to novel methodology development in medical image segmentation in recent years. We reviewed sixteen papers that used the LiTS dataset for method development and evaluation from three sources: a) Journal of Medical Image Analysis (MIA), b) MICCAI conference proceeding, and c) IEEE Transaction on Medical Imaging (TMI), as shown in Table \ref{tab:post_lits}.

A significant advance was on the {3D} deep learning model besides the 2D approaches. \cite{zhou2021models,haghighi2021transferable} proposed self-supervised pre-training frameworks to initialize 3D models for better representation than training them from scratch. \cite{isensee2020nnu} proposed a self-configuring pipeline to facilitate the model training and the automated design of network architecture. \cite{wang2019volumetric} added a 3D attention module for 3D segmentation models. These works improved the efficiency of 3D models and popularized the 3D models in many image segmentation tasks \citep{ma2021cutting}.   

\cite{ma2020learning} focused on the special trait of liver and liver tumor segmentation and proposed a novel active contour-based loss function to preserve the segmentation boundary. Similarly, \cite{tang2020net} proposed to enhance edge information and cross-feature fusion for liver and tumor segmentation. \cite{shirokikh2020universal} considered the varying lesion sizes and proposed a loss reweighting strategy to deal with size imbalance in tumor segmentation. \cite{wang2020conquering} attempted to deal with the heterogeneous image resolution with a multi-branch decoder. 

One emerging trend was leveraging available sparse labeled images to perform multi-organ segmentation.  
\cite{huang2020multi} attempted to perform co-training of single-organ datasets (liver, kidney, and pancreas).  \cite{fang2020multi} proposed a pyramid-input and pyramid-output network to condense multi-scale features to reduce the
semantic gaps. Finally, \cite{yan2020learning} developed a universal lesion detection algorithm to detect a variety of lesions in CT images in a multitask fashion and propose strategies to mine missing annotations from partially-labeled datasets.


\subsubsection{Remaining challenges}
\begin{figure}[!t]
\centering
\includegraphics[width=1\textwidth]{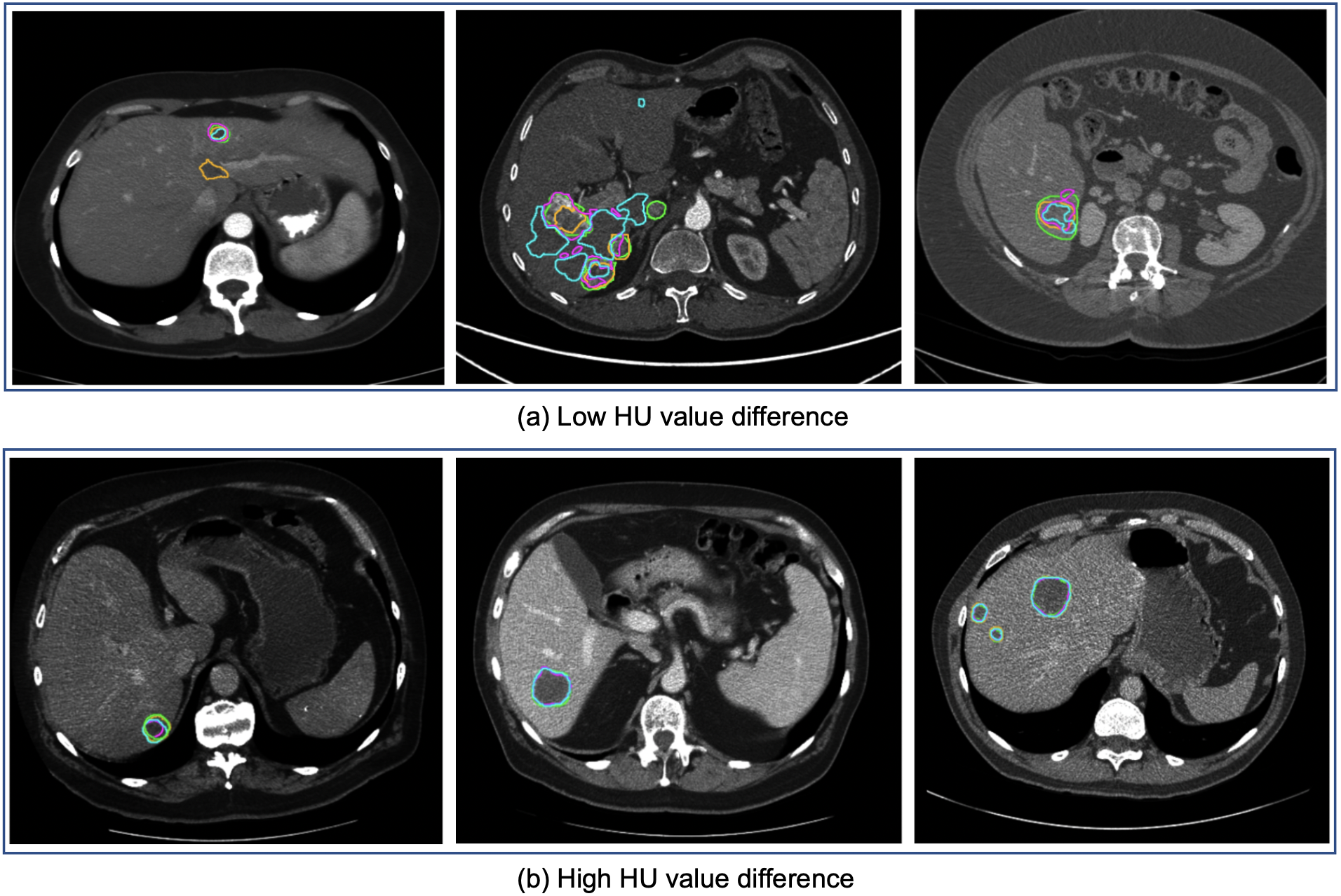}
\caption{Tumor segmentation results with selected cases of the tumor segmentation analysis regarding low ($<$20) and high (40-60) HU value difference. Compared are reference annotation (green), best-performing teams from ISBI 2017 (purple), MICCAI 2017 (orange), and MICCAI 2018 (blue). We can observe that a low HU value difference ($<$20) between tumor and liver tissue poses a challenge for tumor segmentation.}
\label{fig:heatmap_images}
\end{figure}

\paragraph{Segmentation performance w.r.t. lesion size} 
Overall, the submitted methods performed very well for large liver tumors but struggled to segment smaller tumors (see Fig. \ref{fig:heatmap_images}). Many small tumors only have diameters of a few voxels; further, the image resolution is relatively high with 512$\times$512 pixels in axial slices. Therefore, detecting such small structures is difficult due to the small number of potentially differing surrounding pixels, which can indicate a potential tumor border (see Fig. \ref{fig:heatmap_images}). It is exacerbated by the considerable noise and artifacts in medical imaging, which occur from size similarity; texture differences from the surrounding liver tissue and their arbitrary shapes are difficult to distinguish from an actual liver tumor. Overall, state-of-the-art methods performed well on volumes with large tumors and worse on volumes with small tumors. Worst results were achieved in exams where single small tumors ($<$10mm$^{3}$) occur. Best results were achieved when volumes showed less than six tumors with an overall tumor volume above 40mm$^{3}$ (see Fig. \ref{fig:heatmap_images}). 
In the appendix, we show the performance of all submitted methods of the three LiTS challenges, compared for every test volume, clustered by the number of tumor appearances and tumor sizes, see Figure \ref{fig:heatmap}.  

\paragraph{Segmentation performance w.r.t. image contrast} 
Another important influence of the methods' segmentation quality was the difference in tumor and liver HU values. 
Current state-of-the-art methods perform best for volumes showing higher contrast between liver and tumor. Especially in the case of focal lesions with a density 40-60 HU higher than that of the background liver (see Fig. \ref{fig:heatmap_images}). 
Worst results are achieved in cases where the contrast is below 20 HU (see Fig. \ref{fig:heatmap_images}),
including tumors having a lower density than the liver. An average difference in HU values eases the network's task of distinguishing liver and tumor since a simple threshold-derived rule could be applied as part of the decision process. Interestingly, an even more significant difference value did not result in an even better segmentation.

The performance of all submitted methods of three LiTS challenges was compared for every test volume, clustered by the HU level difference between liver and tumor and the HU level difference within tumor ROIs, shown in appendix Figure \ref{fig:heatmaphu}.

\begin{table*}[t]
\scriptsize
\centering
\begin{tabular}{l | l | l}
\hline
\textbf{Source} & \textbf{Authors} & \multicolumn{1}{c}{\textbf{Key Features}} \\ [0.5ex] 

\specialrule{.05em}{-0.1em}{0em}

 MIA & \cite{zhou2021anatomy} &  multimodal registration, unsupervised segmentation, image-guided intervention \\
\cline{2-3}
 MIA & \cite{wang2021pairwise} & 
conjugate fully convolutional network, pairwise segmentation, proxy supervision\\
\cline{2-3}
MIA & \cite{zhou2021models} &   3D Deep learning, self-supervised learning, transfer learning\\
\cline{2-3}
MICCAI & \cite{shirokikh2020universal} & loss reweighting, lesion detection \\
\cline{2-3}
MICCAI & \cite{haghighi2020learning} & self-supervised learning, transfer learning, 3D model pre-training \\ 
\cline{2-3}
MICCAI & \cite{huang2020multi} & co-training of sparse datasets, multi-organ segmentation\\
\cline{2-3}
MICCAI & \cite{wang2019volumetric} & volumetric attention, 3D segmentation \\
\cline{2-3}
MICCAI & \cite{tang2020net} & edge enhanced network, cross feature fusion\\
\cline{2-3}
\tabincell{l}{Nature Methods}   & \cite{isensee2020nnu} & self-configuring framework, extensive evaluation on 23 challenges\\
\cline{2-3}
TMI & \cite{cano2020biomarker} & biomarker regression and localization \\
\cline{2-3}
TMI & \cite{fang2020multi} & multi-organ segmentation, multi-scale training, partially labeled data \\
\cline{2-3}
TMI & \cite{haghighi2021transferable} & self-supervised learning, anatomical visual words\\
 \cline{2-3}
TMI & \cite{zhang2020layer} & interpretable learning, probability calibration\\
\cline{2-3}
TMI & \cite{ma2020learning} & geodesic active contours learning, boundary segmentation \\
\cline{2-3}
TMI & \cite{yan2020learning} & training on partially-labeled dataset, lesion detection, multi-dataset learning\\
\cline{2-3}
TMI & \cite{wang2020conquering} & 2.5D semantic segmentation, attention\\



 \specialrule{.1em}{0em}{-.1em}
 \end{tabular}
\caption{Brief summary of sixteen published work using \emph{LiTS} for the development of novel segmentation methods in medical imaging. While many of them focus on methodological contribution, they also advance the state-of-the-art in liver and liver tumor segmentation.}
\label{tab:post_lits}
\end{table*}


\begin{figure}[!t]
\centering
\includegraphics[width=1\textwidth]{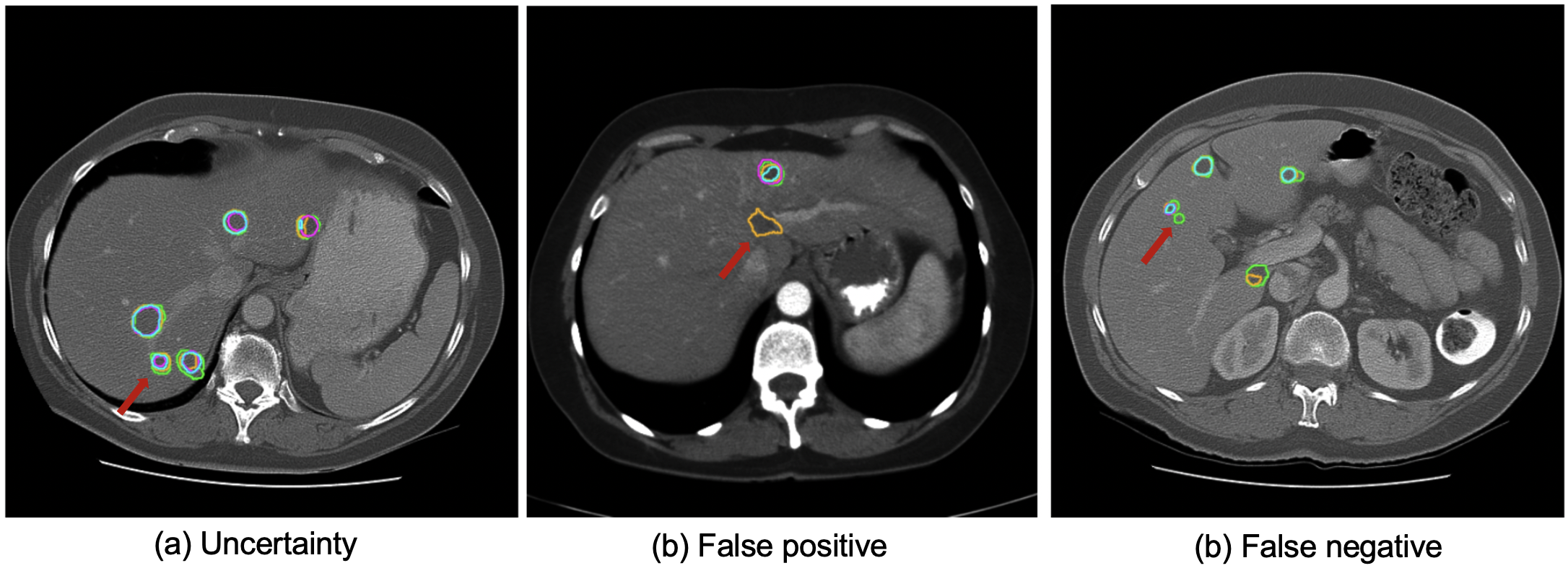}
\caption{Samples of segmentation and detection results for small liver tumor. Compared are reference annotation (green), best-performing teams from ISBI 2017 (purple), MICCAI 2017 (orange), and MICCAI 2018 (blue).}
\label{fig:small_lesion}
\end{figure}

\section{Discussion}
\label{sec:discussion}
\subsection{Limitations}
The datasets were annotated by only one rater from each medical center. Thus it may introduce label bias, especially for small lesion segmentation, which is only ambiguous. However, further quality control of the annotations with consensus can reduce label noise and benefit supervised training and method evaluation. 
 {The initial rankings were conducted considering only the Dice score in which the large tissue will dominate. We observe that solely using Dice does not distinguish the top-performing teams but combining multiple metrics can do better. Unfortunately, imaging information (e.g., scanner type) and demographic information are unavailable when collecting the data from multiple centers. However, they are essential for in-depth analysis and further development of the challenge result \citep{maier2020bias, wiesenfarth2021methods}. The BIAS \citep{maier2020bias} report has proven to provide a good guideline for organizing a challenge and analyzing the outcome of the challenge.} Tumor detection task is of clinical relevance, and the detection metric should be considered in future challenges. 

 To allow for quick evaluation of the submissions, we release the test data to the participant. Hence, we cannot prevent potential overfitting behavior by multiple iterative submissions or cheating behavior (e.g., manual correction of the segmentation). One option to improve this is using image containers (such as Docker \footnote{\url{https://www.docker.com/}} and Singularity \footnote{\url{https://sylabs.io/singularity/}}) without releasing the test images. However, this would potentially limit the popularity of the challenge.
 
\clearpage

\subsection{Future work}
Organizing LiTS has taught us lessons relevant for future medical segmentation benchmark challenges and their organizers. Given that many of the algorithms in this study offered good liver segmentation results compared to tumors, it seems valuable to evaluate liver tumor segmentation based on their different size, type, and occurrence per volume. 
Generating large labeled datasets is time-consuming and costly. It might be more efficiently performed by advanced semi-automated methods, thereby helping to bridge the gap to a fully automated solution.  

Further, we recommend providing multiple reference annotations of liver tumors from multiple raters. This is because the segmentation of liver tumors presents high uncertainty due to the small structure and the ambiguous boundary \citep{schoppe2020deep}. While most of the segmentation tasks in existing benchmarks are formulated to be one-to-one mapping problems, it does not fully solve the image segmentation problem where the data uncertainty naturally exists. Modeling the uncertainty in segmentation task is a trend \footnote{\url{https://qubiq.grand-challenge.org/Home/}} \citep{mehta2020uncertainty,zhang2020disentangling} and would allow the model generates not only one but various plausible outputs. Thus, it would enhance the applicability of automated methods in clinical practice. The released annotated dataset is not limited to benchmarking segmentation tasks but could also serve as data for recent shape modeling methods such as implicit neural functions \citep{yang2022implicitatlas,kuang2022makes,amiranashvili2021learning}
Considering the size and, importantly, the demographic diversity of the patient populations from the seven institutions that contributed cases in the LiTS benchmark dataset, we think its value and contribution to medical image analysis will be greatly appreciated across numerous directions. One example use case is within the research direction of domain adaptation, where the LiTS dataset can be used to account for the apparent shift of the data distribution due to the domain change (e.g., acquisition setting) \citep{glocker2019machine,castro2020causality}. Another recent and intriguing use case is the research direction of federated learning, where the multi-institutional nature of the LiTS benchmark dataset could further contribute to federated learning simulations studies and benchmarks \citep{sheller2018multi,sheller2020federated,rieke2020future,pati2021federated}. It will target potential solutions to the LiTS-related tasks without sharing patient data across institutions. We consider federated learning of particular importance, as scientific maturity in this field could lead to a paradigm shift for multi-institutional collaborations. Furthermore, it is overcoming technical, legal, and cultural data sharing concerns since the patient involved in such collaboration will always be retained within their acquiring institutions.

\section*{Acknowledgement}
Bjoern Menze is supported through the DFG funding (SFB 824, subproject B12) and a Helmut-Horten-Professorship for Biomedical Informatics by the Helmut-Horten-Foundation. Florian Kofler is Supported by Deutsche Forschungsgemeinschaft (DFG) through TUM International Graduate School of Science and Engineering (IGSSE), GSC 81. An Tang was supported by the Fonds de recherche du Qu\'ebec en Sant\'e and Fondation de l'association des radiologistes du Qu\'ebec (FRQS-ARQ 34939 Clinical Research Scholarship – Junior 2 Salary Award).  {Hongwei Bran Li is supported by Forschungskredit (Grant NO.~FK-21-125) from University of Zurich. We thank the CodaLab team, especially Eric Carmichael and Flavio Alexander for helping us with the setup.}
\label{sec:summary}

\clearpage
\bibliographystyle{./01_paper/model5-names}
\bibliography{./refs.bib}


\newpage
\appendix

\section*{ {CRediT Author Statement}}
\noindent
 {Partick Christ: Conceptualization, Writing Original Draft, Data Curation, Visualization, Software, Validation, Investigation. \\
Bjoern Menze: Conceptualization, Writing - Review \& Editing, Data Curation, Visualization, Supervision, Investigation. \\
Partick Bilic: Writing, Data Curation, Visualization, Investigation. \\ 
Hongwei Bran Li: Writing, Revision, Corresponding, Visualization, Investigation.  \\ 
Eugene Vorontsov: Writing, Visualization, Software, Investigation. \\
Rickmer~Braren, Georgios~Kaissis, Avi Ben-Cohen, Adi~Szeskin, Colin Jacobs, Gabriel~Efrain~Humpire~Mamani, Gabriel~Chartrand, Fabian~Loh\"ofer, Julian~Holch, Wieland~Sommer, Felix~Hofmann, Alexandre~Hostettler, Naama~Lev-Cohain, Michal~Drozdzal, Michal Marianne Amitai, Refael~Vivanti, Jacob~Sosna, Volker Heinemann, Christopher Pal, An~Tang, Samuel~Kadoury, Luc~Soler, Bram~van~Ginneken, Hayit~Greenspan: Conceptualization, Data Curation, Writing - Reviewing \& Editing  \\
Ivan~Ezhov, Anjany~Sekuboyina, Fernando~Navarro, Florian~Kofler, Johannes~C.~Paetzold, Suprosanna Shit, Xiaobin~Hu, Jana~Lipkov\'a, Markus~Rempfler, Marie~Piraud, Jan~Kirschke, Benedikt~Wiestler, Zhiheng~Zhang, Christian~H\"ulsemeyer, Marcel~Beetz, Florian~Ettlinger, Michela~Antonelli, Woong~Bae, M\'iriam~Bellver, Lei~Bi, Hao Chen, Grzegorz~Chlebus, Erik~B.~Dam, Qi~Dou, Chi-Wing~Fu, Bogdan~Georgescu, Xavier~Gir\'o-i-Nieto, Felix~Gruen, Xu~Han, Pheng-Ann~Heng, J\"urgen~Hesser, Jan~Hendrik~Moltz, Christian~Igel, Fabian~Isensee, Paul J\"{a}ger, Fucang~Jia, Krishna~Chaitanya~Kaluva, Mahendra~Khened, Ildoo~Kim, Jae-Hun~Kim, Sungwoong~Kim,Simon~Kohl, Tomasz~Konopczynski, Avinash~Kori, Ganapathy~Krishnamurthi, Fan~Li, Hongchao~Li, Junbo~Li, Xiaomeng Li, John Lowengrub, Jun Ma, Klaus~Maier-Hein, Kevis-Kokitsi Maninis, Hans Meine, Dorit Merhof, Akshay Pai, Mathias Perslev, Jens~Petersen, Jordi Pont-Tuset, Jin~Qi, Xiaojuan Qi, Oliver~Rippel, Karsten~Roth, Ignacio~Sarasua, Andrea Schenk, Zengming Shen, Jordi Torres, Christian~Wachinger, Chunliang~Wang, Leon Weninger, Jianrong~Wu, Daguang~Xu, Xiaoping Yang, Simon Chun-Ho Yu, Yading Yuan, Miao~Yue, Liping Zhang: Writing - Review \& Editing.
} \\









\section{Segmentation performance w.r.t tumor size and number of tumors.}
\begin{figure}[H]
\centering
\hspace*{-0.4in}
\includegraphics[width=0.75\textwidth]{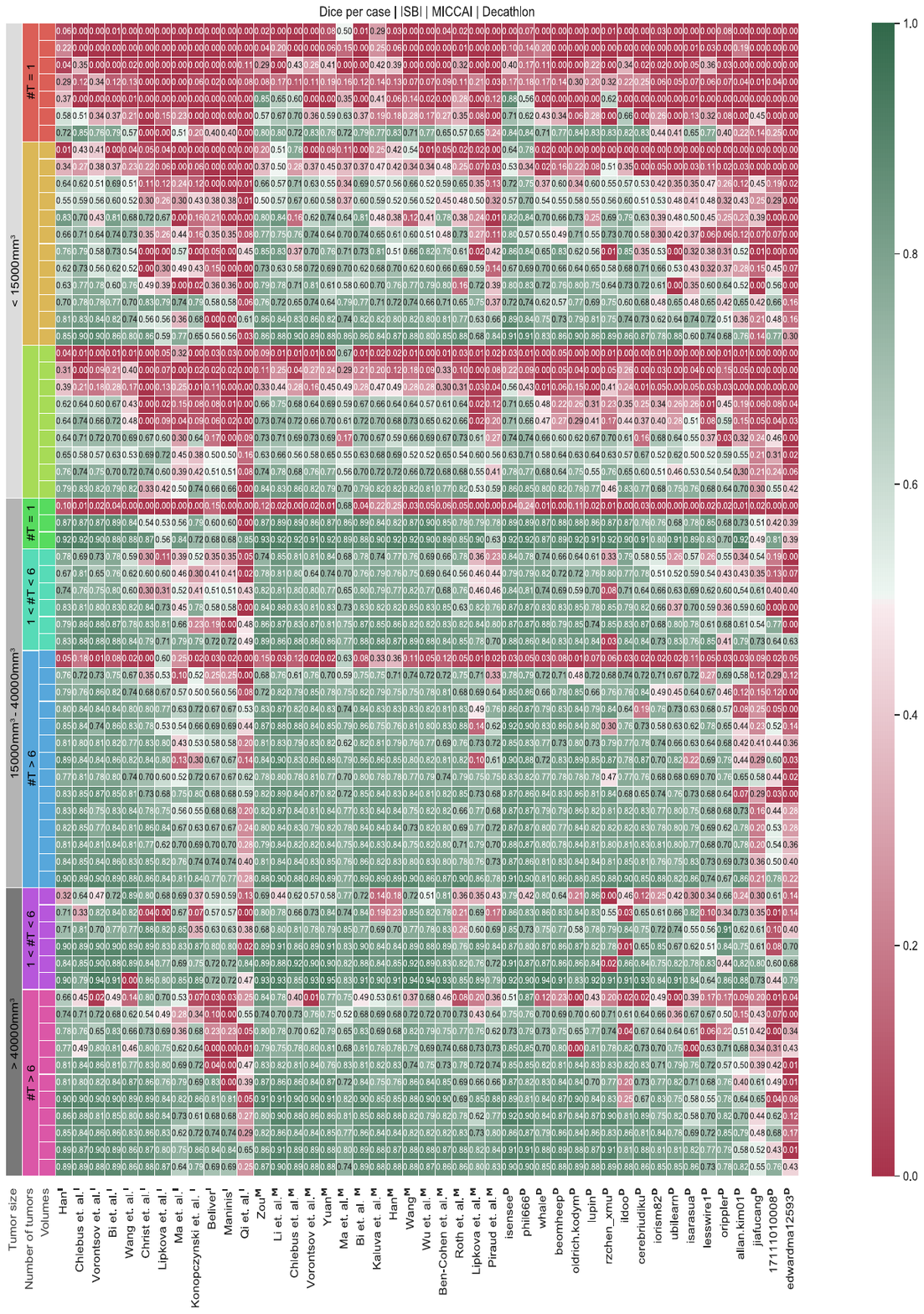}
\caption{Performance w.r.t tumor size and number of tumors. The test dataset is clustered by the number of tumors (\#T) and size of the largest tumor per volume. Overall, participating methods perform well on volumes with large tumors and worse for volumes with small tumors. Worst results are achieved in chase where single small tumors ($<$15mm$^{3}$) occur. Best results are achieved when volumes show less than 6 tumors with an overall tumor volume above 40mm$^{3}$. 
}
\label{fig:heatmap}
\end{figure}

\section{Segmentation performance w.r.t. HU value differences.}
\noindent
\begin{figure}[H]
\centering
\noindent
\includegraphics[width=0.75\textwidth]{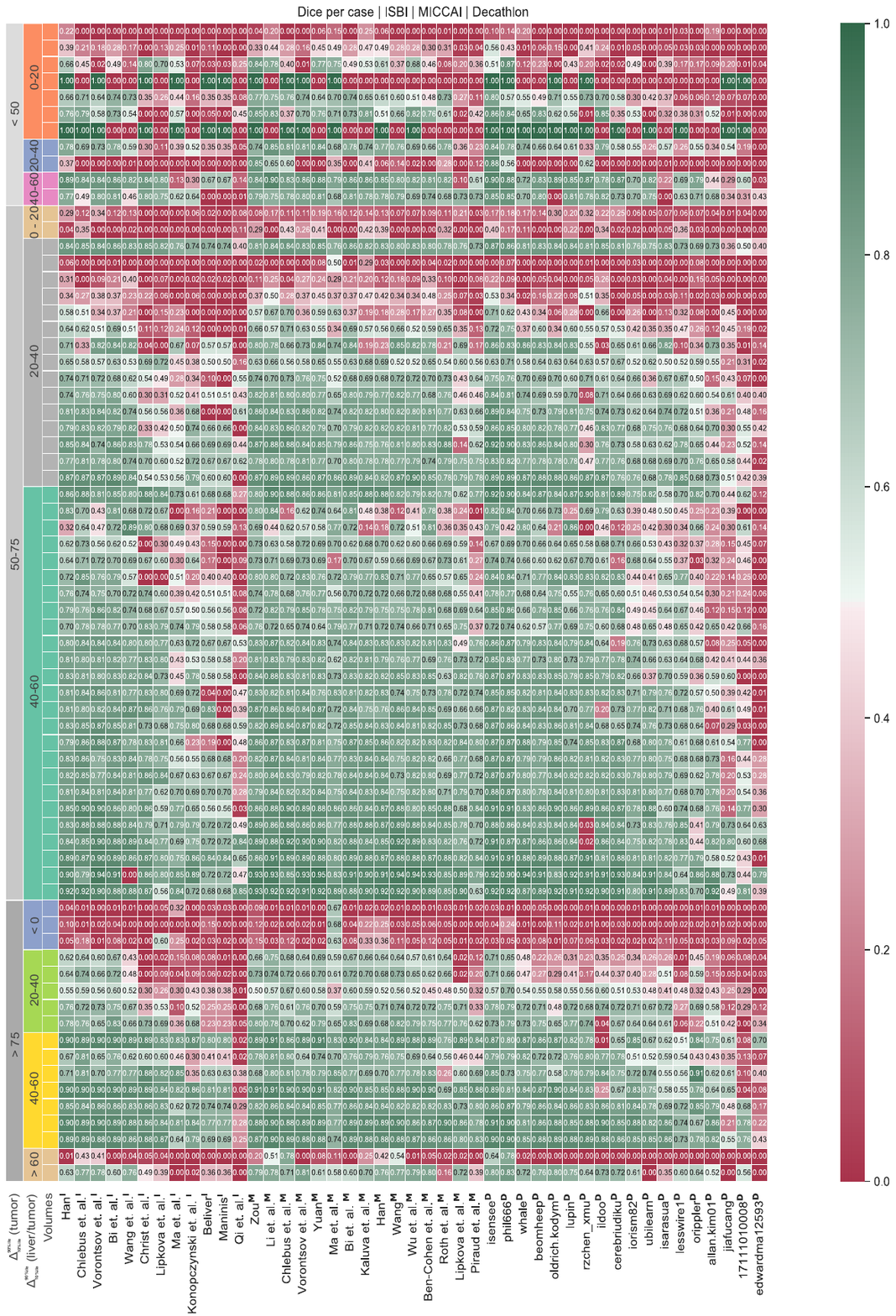}
\vspace{-0.24cm}
\caption{Performance w.r.t HU value difference between tumor and non-tumor liver tissue. Two robust metrics are calculated to cluster the results on the test set. First, the HU value difference between liver and tumor is calculated using both regions' robust median absolute deviation per volume. Further, the clusters are split up by the tumor HU value difference calculated by the difference of the 90$^{th}$ percentile and 10$^{th}$ percentile. Participating methods perform best for volumes showing higher contrast between liver and tumor. Especially in the case of the liver, HU values are 40-60 points higher than the liver. Worst results are achieved in cases where the contrast is below 20 HU value, including tumors having a lower HU value than the liver.  
}
\label{fig:heatmaphu}
\end{figure}


\section{Correspondence algorithm}
\label{correspondence_alg}
\begin{figure}
    \centering
    \includegraphics[width=0.80\linewidth]{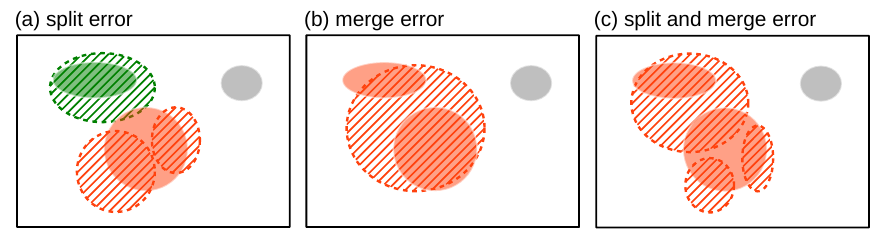}
    \caption{Split and merge errors where a prediction splits a reference lesion into more than one connected component or merges multiple reference components into one, respectively. Reference connected components are shown with a solid color and predicted as regions with a dashed boundary and hatched interior. One-to-one correspondence is shown in green. One-to-two (a), two-to-one (b), and two-to-three (c) correspondence in orange. False negative in gray.}
    \label{fig:split_merge_error}
\end{figure}

Components may not necessarily have a one-to-one correspondence between the two masks. For example, a single reference component can be predicted as multiple components (split error); similarly, multiple reference components can be covered by a single significant predicted component (merge error), as shown in Figure~\ref{fig:split_merge_error}.

\begin{figure}
    \centering
    \includegraphics[width=0.5\linewidth]{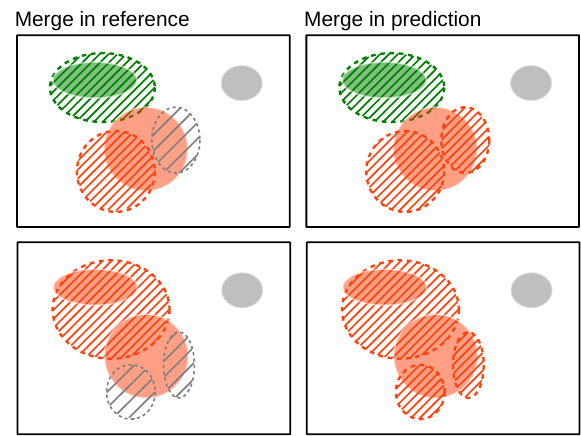}
    \caption{Two examples (top and bottom) of the process to establish a correspondence between connected components in the reference and prediction masks. Reference: solid color; prediction: dashed boundary and hatched interior. Left: reference components merged if the same predicted component overlaps them. Right: predicted components are merged together if the same merged reference component overlaps them. Corresponding reference and predicted components share the same color (green, orange). An undetected reference component is shown in solid gray. During the merge of reference components (left), predicted components that do not have the most significant overlap with a reference component are left unmatched (gray, dashed, and hatched). Their mapping is completed during the merge of predicted components (right).}

    \label{fig:correspondence}
\end{figure}

Once connected components are found and enumerated in the reference and prediction masks, the correspondence algorithm determines the mapping between reference and prediction. Consider $N_R$ connected components in the reference mask and $N_P$ in the prediction mask. First, the many-to-many mapping is turned into a many-to-one mapping by merging all reference components $\mathbf{r_i}$ ($i \in N_R$) that are connected by a single predicted component $\mathbf{p_j}$ ($j \in N_P$), as shown in Figure~\ref{fig:correspondence} (left). In the case where an $\mathbf{r_i}$ overlaps multiple $\mathbf{p_j}$, the $\mathbf{p_j}$ with the largest total intersected area is used. Thus, for every index $i \in N_R$, a corresponding $j_i \in N_P$ is determined as: 

\begin{equation}
    \forall i \in N_R, ~j_i = 
    \begin{cases}
        \argmax_j \mathbf{p_j} \cap \mathbf{r_i}, & \mathbf{p_j} \cap \mathbf{r_i} > 0 \\
        \text{none}, & \text{else}
    \end{cases}
    \label{eq:correspondence_algorithm_step1_match}
\end{equation}



and the $\mathbf{r_i}$ are merged to $\mathbf{\rho_j}$ according to:

\begin{equation}
    \mathbf{\rho_j} = \bigcap_{i : j_i=j} \mathbf{r_i},
    \label{eq:correspondence_algorithm_step1_merge}
\end{equation}

resulting in $N_\rho = |\{\mathbf{\rho_j}\}|$ regions. Any $\mathbf{r_i}$ without a corresponding $\mathbf{p_j}$ is a false negative.

In the second step, the mapping is completed by associating each remaining $\mathbf{p_j}$ with a single $\mathbf{\rho_k}$ ($k \in N_\rho$) with which it shares the largest total intersected area and merging all the $\mathbf{p_j}$ that share the same $\mathbf{\rho_k}$, as shown in Figure~\ref{fig:correspondence} (right). Thus, for every index $j \in N_P$, a corresponding $k \in N_\rho$ is determined as:

\begin{equation}
    \forall j \in N_P, ~k_j = 
    \begin{cases}
        \argmax_k \mathbf{p_j} \cap \mathbf{\rho_k}, & \mathbf{p_j} \cap \mathbf{\rho_k} > 0 \\
        \text{none}, & \text{else}
    \end{cases}
    \label{eq:correspondence_algorithm_step2_match}
\end{equation}

and the $\mathbf{p_j}$ are merged to $\mathbf{\pi_k}$ according to:

\begin{equation}
    \mathbf{\pi_k} = \bigcap_{j : k_j=k} \mathbf{p_j},
    \label{eq:correspondence_algorithm_step1_merge}
\end{equation}

resulting in $N_\pi = |\{\mathbf{\pi_k}\}|$ regions. Any $\mathbf{p_j}$ without a corresponding $\mathbf{r_i}$ is a false positive.

The result is a map with a correspondence between sets of predicted components and sets of reference components. In order to maintain the immutability of the reference, any metrics evaluated on a set of merged reference components are attributed to each constituent reference component. For example, if two connected components in the reference mask are merged, and a Dice score of 0.7 is computed on the mask combining both components, then each component is considered to have a Dice score of 0.7. If the reference components were not merged, the errors computed for each of the two components would be exaggerated (e.g., 0.3 and 0.5 Dice).

\section{Automated tumor burden analysis in MICCAI-LiTS 2017}
\vspace{-0.1cm}
\paragraph{Motivation} The tumor burden, defined as the liver/tumor ratio, plays an essential role in surgical resection planning \citep{nordlinger1996surgical, jagannath1986tumor}. Instead of measuring diameters of target tumors, a fully volumetric segmentation of both the liver and its tumor and the subsequent tumor burden analysis offers valuable insights into the disease progression \citep{blachier2013burden}. Further, tumor burden is also essential in assessing the effectiveness of different treatments and can potentially replace the RECIST protocol \citep{gobbi2004clinical, bornemann2007oncotreat, heussel2007follow, kuhnigk2006morphological, puesken2010prediction,bauknecht2010intra}. A fully automated liver and tumor segmentation allows more straightforward computation of tumor burden and simplifies surgical liver resection planning.

\paragraph{Metrics}
The tumor burden of the liver is a measure of the fraction of the liver afflicted by cancer. As a metric, we measure the root mean square error (RMSE) in tumor burden estimates from lesion predictions.
\begin{equation}\label{eq:rmse}
RMSE = \sqrt{\frac{1}{n}\sum_{i=1}^{n}{\Big(A_i -B_i}\Big)^2}
\end{equation}

\begin{table}[]
\scriptsize
\centering
\begin{tabular}{@{}llllll@{}}
\toprule
Ranking & Ref. Name                  & Institution       & RMSE          & Max Error   \\ \toprule
1       & C. Li \emph{et al.}       & CUHK              & 0.015 (1)    & 0.062 (6)  \\
2       & J. Wu \emph{et al.}        & NJU               & 0.016 (2)    & 0.048 (2)  \\
3       & C. Wang~\emph{et al.}     & KTH               & 0.016 (3)    & 0.058 (4)  \\
4       & Y. Yuan \emph{et al.}    & MSSM              & 0.017 (4)    & 0.049 (3)  \\
5       & J. Zou \emph{et al.}  & Lenovo            & 0.017 (5)        & 0.045 (1)  \\
6       & K. Kaluva \emph{et al.}    & Predible Health   & 0.020 (6)    & 0.090 (12) \\
7       & X. Han~\emph{et al.}   & Elekta Inc.       & 0.020 (7)        & 0.080 (10) \\
8       & A. Ben-Cohen \emph{et al.}      & Uni Tel Aviv & 0.020 (8)    & 0.070 (7) \\
9       & G. Chlebus \emph{et al.}     & Fraunhofer        & 0.020 (9)    & 0.070 (8) \\
10    & L. Zhang \emph{et al.}     & CUHK    & 0.022 (10)  & 0.074 (11) \\
11       & E. Vorontsov \emph{et al.}    & MILA             & 0.023 (11)   & 0.112 (13) \\
12       & J. Lipkova \emph{et al.}    & TUM               & 0.030 (12)   & 0.140 (14) \\
13       & K. Roth \emph{et al.}        & Volume Graphics   & 0.030 (13)   & 0.180 (15) \\
14       & M. Piraud \emph{et al.}   & TUM               & 0.037 (14)         & 0.143 (16) \\
15   & Jin Qi           & 0.0420 (12) & 0.0330 (2)  \\
16      & L. Bi \emph{et al.}      & Uni Sydney             & 0.170 (15)   & 0.074 (9) \\
17      & J. Ma \emph{et al.}   & NJUST                  & 0.920 (16)   & 0.061 (5)  \\
\bottomrule
\end{tabular}%
\caption{Tumor burden ranking in MICCAI-LiTS 2017.}
\label{tab:results_tumor_burden}
\end{table}
\vspace{-0.1cm}
\paragraph{Results} The tumor burden was well predicted by many methods, with the best-performing method achieving the lowest RMSE of 0.015 and the lowest maximum error at 0.033 (Tab. ~\ref{tab:results_tumor_burden}). There was a slight variation in RMSE values by the last. 15ˆ{th} method still obtains the fifth rank due to the high number of duplicates in the low range of values. In overall, methods achieving high Dice per case scores also obtained lower RMSE values. Only small correlation exists between RMSE and the maximum error ranking.

\end{document}